\documentclass{article} 
\usepackage{iclr2020_conference,times}

\usepackage{algorithm}
\usepackage{algorithmic}
\usepackage{bbm}

\usepackage{amsmath,amsfonts,bm}

\def\eqref#1{equation~\ref{#1}}
\def\Eqref#1{Eq.~\ref{#1}}

\def\1{\bm{1}}

\def\eps{{\epsilon}}

\DeclareMathAlphabet{\mathsfit}{\encodingdefault}{\sfdefault}{m}{sl}
\SetMathAlphabet{\mathsfit}{bold}{\encodingdefault}{\sfdefault}{bx}{n}

\DeclareMathOperator*{\argmax}{arg\,max}
\DeclareMathOperator*{\argmin}{arg\,min}

\usepackage{url}
\usepackage{booktabs}
\usepackage{nicefrac}
\usepackage{overpic}
\usepackage{graphicx}
\usepackage{microtype}                                                          
\usepackage{graphicx}                    
\usepackage{siunitx}
\usepackage{times}
\usepackage{subcaption}
\usepackage{colortbl}
\usepackage{color}
\usepackage{multirow}
\usepackage{enumitem} 

\usepackage{pifont}
\definecolor{ourmethod}{gray}{0.93}
\definecolor{priormethod}{gray}{1} 
\usepackage[utf8]{inputenc} 
\usepackage[T1]{fontenc}    
\usepackage{hyperref}       
\usepackage{url}            
\usepackage{booktabs}       
\usepackage{amsfonts}       
\usepackage{nicefrac}       
\usepackage{microtype}      

\usepackage{mathrsfs}
\DeclareMathAlphabet{\mathpzc}{OT1}{pzc}{m}{it}

\usepackage[bottom,hang,flushmargin]{footmisc}

\definecolor{mydarkblue}{rgb}{0,0.08,0.45}
\hypersetup{ %
    pdftitle={},
    pdfauthor={},
    pdfsubject={},
    pdfkeywords={},
    pdfborder=0 0 0,
    pdfpagemode=UseNone,
    colorlinks=true,
    linkcolor=mydarkblue,
    citecolor=mydarkblue,
    filecolor=mydarkblue,
    urlcolor=mydarkblue,
    pdfview=FitH}

\iclrfinalcopy

\definecolor{myredcolor}{RGB}{215,48,39}
\definecolor{mygreencolor}{RGB}{26,152,80}
\newcommand{\cmark}{\textcolor{mygreencolor}{\ding{51}}}
\newcommand{\xmark}{\textcolor{myredcolor}{\ding{55}}}

\newcommand\FBox[1]{{\setlength{\fboxsep}{-0.5pt}\setlength{\fboxrule}{0.5pt}\frame{#1}}}

\newcolumntype{D}{>{\setbox0=\hbox\bgroup}c<{\egroup}@{}}

\newcommand{\latent}[0]{\mathbf{z}} 
\newcommand{\Latent}[0]{\mathbf{Z}} 
\newcommand{\State}[0]{\mathbf{S}}
\newcommand{\state}[0]{\mathbf{s}}
\newcommand{\goal}[0]{\mathbf{g}}  
\newcommand{\action}[0]{\mathbf{a}}

\newcommand{\bsigma}{\boldsymbol{\sigma}}

\newcommand{\bchi}{\boldsymbol{\chi}}
\newcommand{\bGamma}{\boldsymbol{\Gamma}}
\newcommand{\blambda}{\boldsymbol{\lambda}}

\newcommand\alltime{}
\newcommand{\deriv}[2]{\frac{\text{d}#1}{\text{d}#2}}

\title{Deep Imitative Models for \\ Flexible Inference, Planning, and Control}

\author{%
 Nicholas Rhinehart \\
 Carnegie Mellon University \\
 {\small  \texttt{nrhineha@cs.cmu.edu} }
  \And
  Rowan McAllister \\
  UC Berkeley \\
 {\small \texttt{rmcallister@berkeley.edu} }
  \And 
  Sergey Levine \\
  UC Berkeley \\
  {\small \texttt{svlevine@berkeley.edu}  }
}

\begin{document}

\maketitle

\begin{abstract}
Imitation Learning (IL) is an appealing approach to learn desirable autonomous behavior. However, \emph{directing} IL to achieve arbitrary goals is difficult.
In contrast, planning-based algorithms use dynamics models and reward functions to achieve goals. Yet, reward functions that evoke desirable behavior are often difficult to specify. 
In this paper, we propose ``Imitative Models'' to combine the benefits of IL and goal-directed planning. Imitative Models are probabilistic predictive models of desirable behavior able to plan interpretable expert-like trajectories to achieve specified goals.
We derive families of flexible goal objectives, including constrained goal regions, unconstrained goal sets, and energy-based goals. We show that our method can use these objectives to successfully direct behavior. Our method substantially outperforms six IL approaches and a planning-based approach in a dynamic simulated autonomous driving task, and is efficiently learned from expert demonstrations without online data collection.  We also show our approach is robust to poorly specified goals, such as goals on the wrong side of the road.
\end{abstract}

\section{Introduction}

Imitation learning (IL) is a framework for learning a model to mimic behavior. At test-time, the model pursues its best-guess of desirable behavior. By letting the model choose its own behavior, we cannot \emph{direct} it to achieve different goals.
While work has augmented IL with goal conditioning~\citep{dosovitskiy2016learning,codevilla2018end}, it requires goals to be specified during training, explicit goal labels, and are simple (e.g., turning). In contrast, we seek flexibility to achieve \emph{general} goals for which we have \emph{no demonstrations}.

In contrast to IL, planning-based algorithms like model-based reinforcement learning (MBRL) methods do not require expert demonstrations. MBRL can adapt to new tasks specified through reward functions~
\citep{kuvayev96model,deisenroth2011pilco}. The ``model'' is a dynamics model, used to \emph{plan} under the user-supplied reward function. Planning enables these approaches to perform new tasks at test-time. 
The key drawback is that these models learn dynamics of \emph{possible} behavior rather than dynamics of \emph{desirable} behavior. This means that the responsibility of evoking desirable behavior is entirely deferred to engineering the input reward function. 
Designing reward functions that cause MBRL to evoke complex, desirable behavior is difficult when the space of possible \emph{undesirable} behaviors is large. 
In order to succeed, the rewards cannot lead the model astray towards observations significantly different than those with which the model was trained.

\begin{figure*}[htpb]
    \centering
    \includegraphics[width=\textwidth]{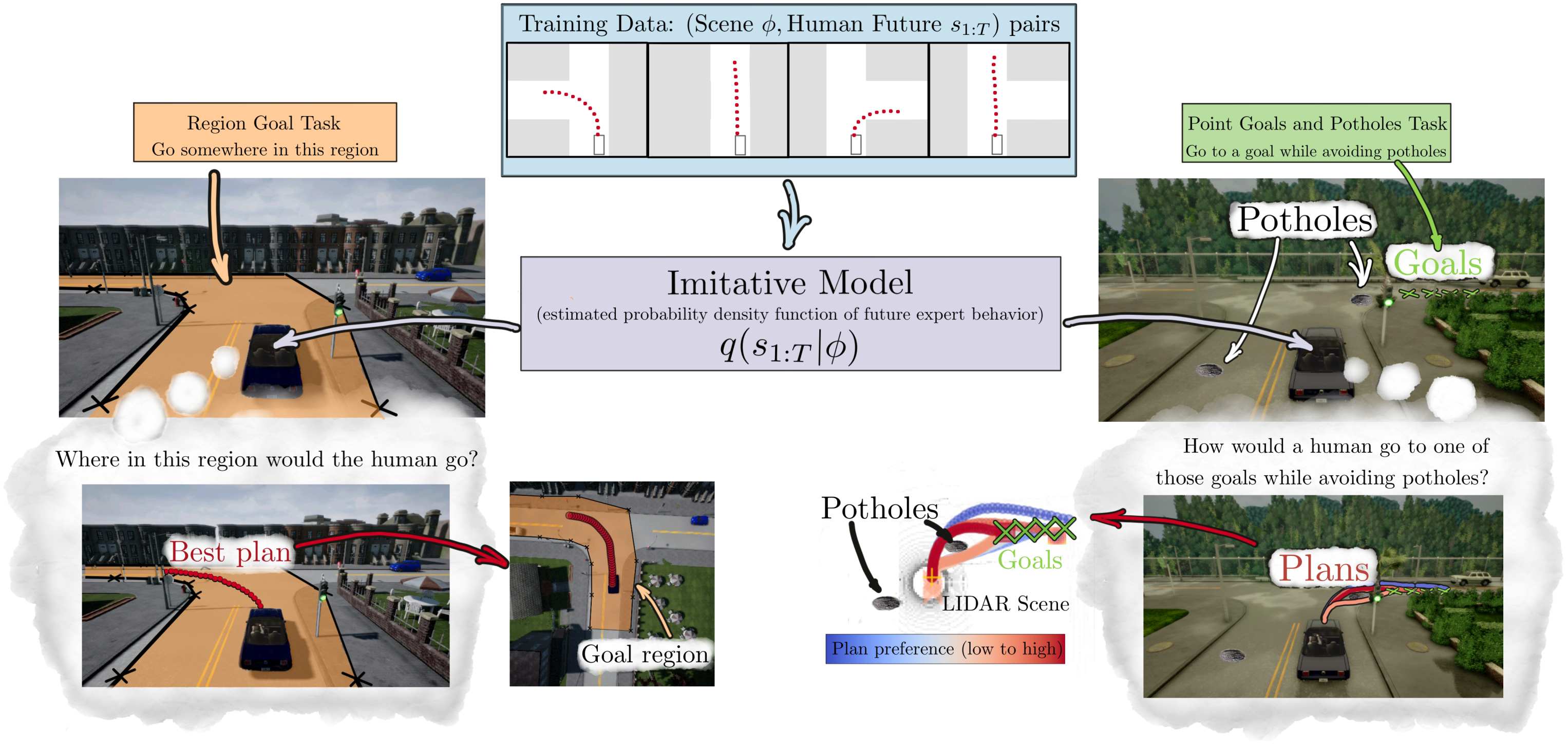}
    \caption{Our method: deep imitative models. \emph{Top Center}. We use demonstrations to learn a probability density function $q$ of future behavior and deploy it to accomplish various tasks. \emph{Left}: A region in the ground plane is input to a planning procedure that reasons about how the expert would achieve that task. It coarsely specifies a destination, and guides the vehicle to turn left. \emph{Right}: Goal positions and potholes yield a plan that avoids potholes and achieves one of the goals on the right.}
    \label{fig:teaser}
\end{figure*}



Our goal is to devise an algorithm that combines the advantages of MBRL and IL by offering MBRL's flexibility to achieve new tasks at test-time and IL's  potential to learn desirable behavior entirely from offline data. To accomplish this, we first train a model to forecast expert trajectories with a density function, which can \emph{score trajectories and plans by how likely they are to come from the expert}. A probabilistic model is necessary because expert behavior is stochastic: e.g. at an intersection, the expert could choose to turn left or right. Next, we derive a principled probabilistic inference objective to create plans that incorporate both (1) the model and (2) arbitrary new tasks. Finally, we derive families of tasks that we can provide to the inference framework. Our method can accomplish \emph{new tasks specified as complex goals} without having seen an expert complete these tasks before.


We investigate properties of our method on a dynamic simulated autonomous driving task (see Fig.~\ref{fig:teaser}). Videos are available at \mbox{\small \url{https://sites.google.com/view/imitative-models}}. Our contributions are as follows:

\begin{enumerate}[leftmargin=*] \itemsep0em
    \item {\bf Interpretable expert-like plans without reward engineering.} Our method outputs multi-step expert-like plans, offering superior interpretability to one-step imitation learning models. In contrast to MBRL, our method generates expert-like behaviors without reward function crafting.
    \item{\bf Flexibility to new tasks:} In contrast to IL, our method flexibly incorporates and achieves goals not seen during training, and performs complex tasks that were never demonstrated, such as navigating to goal regions and avoiding test-time only potholes, as depicted in Fig.~\ref{fig:teaser}.
    \item {\bf Robustness to goal specification noise:} We show that our method is robust to noise in the goal specification. In our application, we show that our agent can receive goals on the wrong side of the road, yet still navigate towards them while staying on the correct side of the road.
    \item{\bf State-of-the-art CARLA performance:}  Our method substantially outperforms MBRL, a custom IL method, and all five prior CARLA IL methods known to us. It learned near-perfect driving through dynamic and static CARLA environments from expert observations alone.
\end{enumerate}

\section{Deep Imitative Models}\label{section:method}

We begin by formalizing assumptions and notation. We model continuous-state, discrete-time, partially-observed Markov processes. Our agent's state at time $t$ is $\state_t \in \mathbb{R}^{D}$; $t=0$ refers to the current time step, and $\phi$ is the agent's observations. Variables are bolded. Random variables are capitalized. Absent subscripts denote \textit{all} future time steps, e.g.\ $\State \doteq \State_{ 1:T } \in \mathbb{R}^{T\times D}$. 
We denote a probability density function of a random variable $\State$ as $p(\State)$, and its value as $p(\state)\doteq p(\State\!=\!\state)$.

To learn agent dynamics that are possible and preferred, we construct a model of expert behavior.
We fit an ``Imitative Model'' $q(\State_{1:T}|\phi)=\prod_{t=1}^T q(\State_t|\State_{1:t-1},\phi)$ to a dataset of expert trajectories $\mathcal D = \{(s^i, \phi^i)\}_{i=1}^N$ drawn from a (unknown) distribution of expert behavior $s^i \sim p(\State|\phi^i)$. By training $q(\State|\phi)$ to forecast expert trajectories with high likelihood, we model the scene-conditioned expert dynamics, which can score trajectories by how likely they are to come from the expert.

\subsection{Imitative Planning to Goals}
After training, $q(\State\alltime|\phi)$ can generate trajectories that resemble those that the expert might generate -- e.g. trajectories that navigate roads with expert-like maneuvers. However, these maneuvers will not have a specific goal. Beyond generating human-like behaviors, we wish to \textit{direct} our agent to goals and have the agent automatically reason about the necessary mid-level details. We define general tasks by a set of goal variables $\mathcal G$. The probability of a plan $\state$ conditioned on the goal $\mathcal G$ is modelled by a posterior  $p(\state\alltime | \mathcal G,  \phi)$. This posterior is implemented with $q(\state\alltime | \phi)$  as a \emph{learned} imitation prior and $p(\mathcal G | \state\alltime, \phi)$ as a \emph{test-time} goal likelihood. We give examples of $p(\mathcal G | \state\alltime, \phi)$ after deriving a maximum a posteriori  inference procedure to generate expert-like plans that achieve abstract goals:
\begin{align}
\state^* \doteq \argmax_{\state\alltime} \; \log p(\state\alltime | \mathcal G,\phi) \nonumber &= \argmax_{\state\alltime} \; \log q(\state\alltime |  \phi) + \log p(\mathcal G | \state\alltime, \phi) - \log p(\mathcal G |  \phi )  \nonumber \\ 
&= \argmax_{\state\alltime} \; \log \!\!\!\! \underbrace{q(\state\alltime |  \phi)}_{\text{imitation prior}} \!\!\!\! + \log \! \underbrace{p(\mathcal G| \state\alltime, \phi)}_{\text{goal likelihood}} \!\! . 
\label{eqn:goal-sequence-planning} 
\end{align}
We perform gradient-based optimization of Eq.~\ref{eqn:goal-sequence-planning}, and defer this discussion to Appendix~\ref{sec:pseudocode}. Next, we discuss several goal likelihoods, which direct the planning in different ways. They communicate \emph{goals} they desire the agent to achieve, but not how to achieve them. The planning procedure determines how to achieve them by producing paths similar to those an expert would have taken to reach the given goal. In contrast to black-box one-step IL that predicts controls, our method produces interpretable \emph{multi-step plans} accompanied by two scores. One estimates the plan's ``expertness'', the second estimates its probability to achieve the goal. Their sum communicates the plan's overall quality.

\subsection{Constructing Goal Likelihoods} \label{sec:goallikelihoods}
\noindent{\bf Constraint-based planning to goal sets (hyperparameter-free)}: Consider the setting where we have access to a set of desired final states, one of which the agent should achieve. We can model this by applying a Dirac-delta distribution on the final state, to ensure it lands in a goal set $\mathbb G\subset\!\mathbb R^D$: 
\begin{align}
    p(\mathcal G|\state\alltime, \phi) 
    \leftarrow \delta_{\state_T}(\mathbb G), \quad  \delta_{\state_T}(\mathbb G)=1~\text{if}\;\; \state_T \in \mathbb G, \quad \delta_{\state_T}(\mathbb G)=0~\text{if}\;\; \state_T \not\in \mathbb G.
\end{align}
$\delta_{\state_T}(\mathbb G)$'s \emph{partial support} of $\state_T\!\in \mathbb G\subset\!\mathbb R^D$ constrains $\state_T$ and introduces \emph{no hyperparameters} into $p(\mathcal G|\state\alltime, \phi)$.  For each choice of $\mathbb G$, we have a different way to provide high-level task information to the agent. The simplest choice for $\mathbb G$ is a \emph{finite} set of points: a {\bf (A) Final-State Indicator} likelihood. We applied {(A)} to a sequence of \emph{waypoints} received from a standard A$^*$ planner (provided by the CARLA simulator), and outperformed all prior dynamic-world CARLA methods known to us. We can also consider providing an \emph{infinite} number of points. Providing a set of line-segments as $\mathbb G$ yields a {\bf (B) Line-Segment Final-State Indicator} likelihood, which encourages the final state to land along one of the segments. Finally, consider a {\bf (C) Region Final-State Indicator} likelihood in which $\mathbb G$ is a polygon (see Figs.~\ref{fig:teaser}~and~\ref{fig:single_bigregion}).
Solving Eq.~\ref{eqn:goal-sequence-planning} with (C) amounts to \emph{planning the most expert-like trajectory that ends inside a goal region}. Appendix~\ref{app:goaldetails} provides derivations, implementation details, and additional visualizations. We found these methods to work well when $\mathbb G$ contains ``expert-like'' final position(s), as the prior strongly penalizes plans ending in non-expert-like positions. 


\noindent{\bf Unconstrained planning  to goal sets (hyperparameter-based)}: Instead of \emph{constraining} that the final state of the trajectory reach a goal, we can use a goal likelihood with \emph{full support} ($\state_T\!\in\!\mathbb R^D$), centered at a desired final state. \emph{This lets the goal likelihood encourage goals, rather than dictate them.} If there is a single desired goal ($\mathbb G\!=\!\{\goal_T\}$), the {\bf (D) Gaussian Final-State} likelihood $p(\mathcal G|\state\alltime, \phi) \leftarrow \mathcal N(\goal_T;\state_T, \eps I)$ treats  $\goal_T$ as a \emph{noisy} observation of a final future state, and encourages the plan to arrive at a final state.
We can also plan to $K$ successive states $\mathcal G = (\goal_{T-K+1}, \dots, \goal_{T})$ with a  {\bf (E) Gaussian State Sequence:} \smash{$ p(\mathcal G|\state\alltime, \phi) \leftarrow \prod_{k=T-K+1}^{T} \mathcal N(\goal_{k};\state_{k}, \eps I)$} if a program wishes to specify a desired end velocity or acceleration when reaching the final state $\goal_T$ (Fig.~\ref{fig:sequence_planning}). Alternatively, a planner may propose a set of states with the intention that the agent should reach any one of them. This is possible by using a {\bf (F) Gaussian Final-State Mixture:} \smash{$ p(\mathcal G|\state\alltime, \phi) \leftarrow {\scriptsize \frac{1}{K}}\sum_{k=1}^{K} \mathcal N(\goal^{k}_{T};\state_{T}, \eps I)$} and is useful if some of those final states are not reachable with an expert-like plan.
Unlike A--C, D--F introduce a hyperparameter ``$\epsilon$''. However, they are useful when \emph{no states in $\mathbb G$ correspond to observed expert behavior}, as they allow the imitation prior to be robust to poorly specified goals.
 
\noindent {\bf Costed planning}: Our model has the additional flexibility to accept arbitrary user-specified costs $c$ at test-time. For example, we may have updated knowledge of new hazards at test-time, such as a given map of potholes or a predicted cost map. Cost-based knowledge $c(\state_i|\phi)$ can be incorporated as an {\bf (G) Energy-based} likelihood: $p(\mathcal G| \state\alltime, \phi) \propto \prod_{t=1}^T e^{-c(\state_t|\phi)}$ \citep{todorov2007linearly,levine2018reinforcement}. This can be combined with other goal-seeking objectives by simply \emph{multiplying} the likelihoods together. Examples of combining G (energy-based) with F (Gaussian mixture) were shown in Fig.~\ref{fig:teaser} and are shown in Fig.~\ref{fig:pothole_planning}. Next, we describe  instantiating $q(\State|\phi)$ in CARLA \citep{dosovitskiy17carla}.    

\newlength{\pplanlength}
\setlength{\pplanlength}{.181\columnwidth}
\newlength{\pplanfboxlength}
\setlength{\pplanfboxlength}{.18\columnwidth}
\begin{figure}[hbt]
    \centering
    \begin{minipage}[t]{.38\textwidth}
        \centering
           \begin{subfigure}[t]{\pplanlength}
           \includegraphics[width=\textwidth]{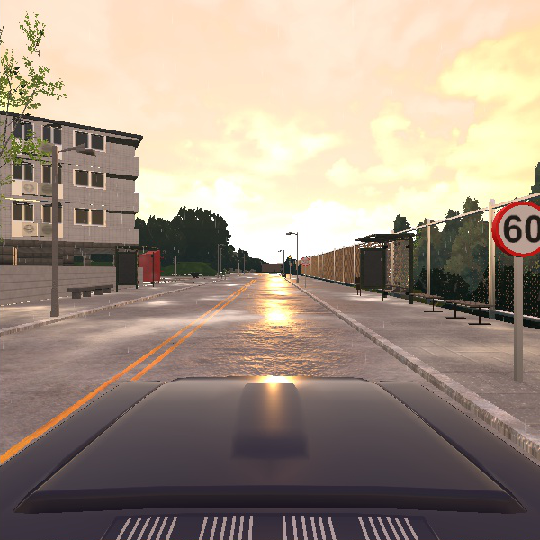}
        \end{subfigure}
         \FBox{    \begin{subfigure}[t]{\pplanfboxlength}
           \begin{overpic}[width=\textwidth,angle=90]{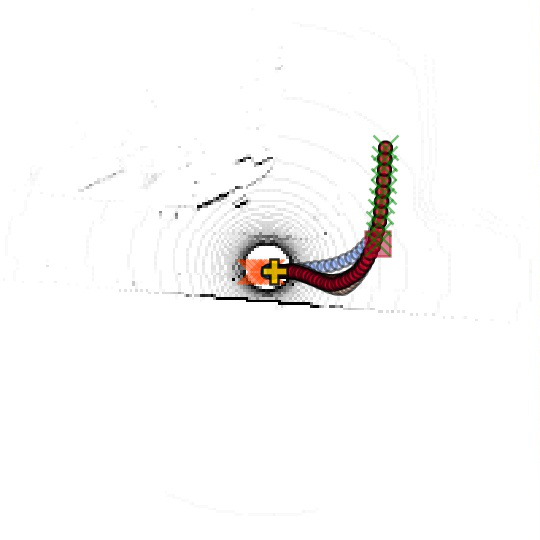}
            \put(0,15){\includegraphics[width=0.7\textwidth]{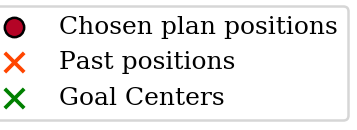}}
           \put(0,0){\includegraphics[width=0.85\textwidth]{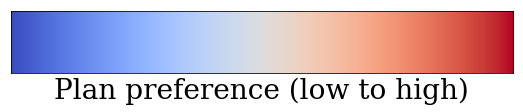}}
           \end{overpic} 
           \end{subfigure}}
    \caption{Imitative planning with the Gaussian State Sequence enables fine-grained control of the plans.} \label{fig:sequence_planning}
    \end{minipage}%
    \hfill
    \begin{minipage}[t]{0.39\textwidth}
    \centering
        \begin{subfigure}[t]{\pplanlength}
             \includegraphics[width=\textwidth]{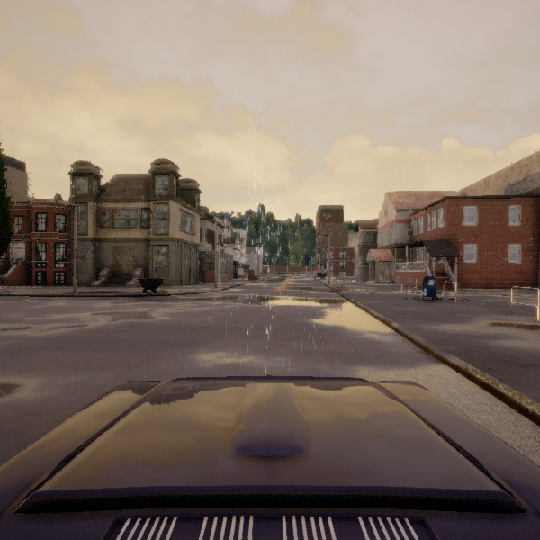}
        \end{subfigure} 
         \begin{subfigure}[t]{\pplanfboxlength}
         \FBox{     \begin{overpic}[width=\textwidth,angle=90]{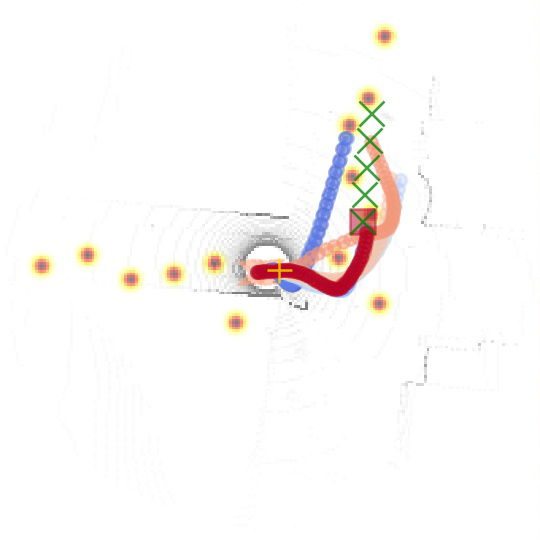}
                 \put(0,15){\includegraphics[width=0.7\textwidth]{img/legends/teaser_plotlegend.png}}
             \put(0,0){\includegraphics[width=0.85\textwidth]{img/legends/coolwarm_lth.png}}
             \end{overpic} }
         \end{subfigure} 
              \caption{Costs can be assigned to ``potholes'' only seen at test-time. The planner prefers routes avoiding potholes. 
              } 
              \label{fig:pothole_planning}
     \end{minipage}
     \hfill
     \begin{minipage}[t]{0.19\textwidth}
    \centering
      \begin{subfigure}[t]{\textwidth}
    \includegraphics[angle=90,width=\textwidth]{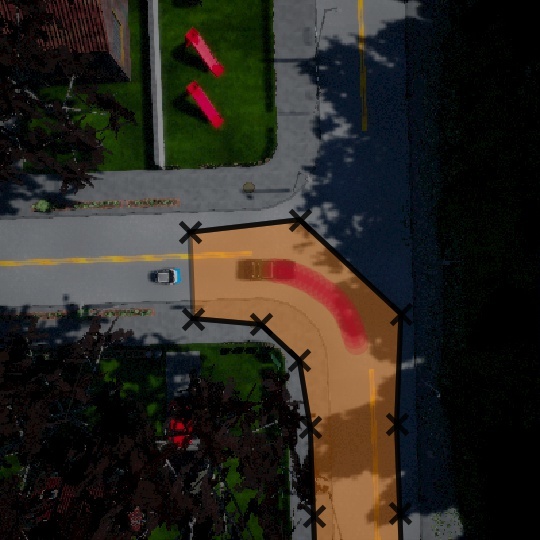}
    \end{subfigure} 
    \caption{Goal regions can be coarsely specified to give directions.} \label{fig:single_bigregion}
    \end{minipage}
\end{figure}

\subsection{Applying Deep Imitative Models to Autonomous Driving}  \label{sec:implementation}
\begin{figure}[htb]
\centering
\begin{subfigure}[t]{0.85\textwidth}
\includegraphics[width=\textwidth]{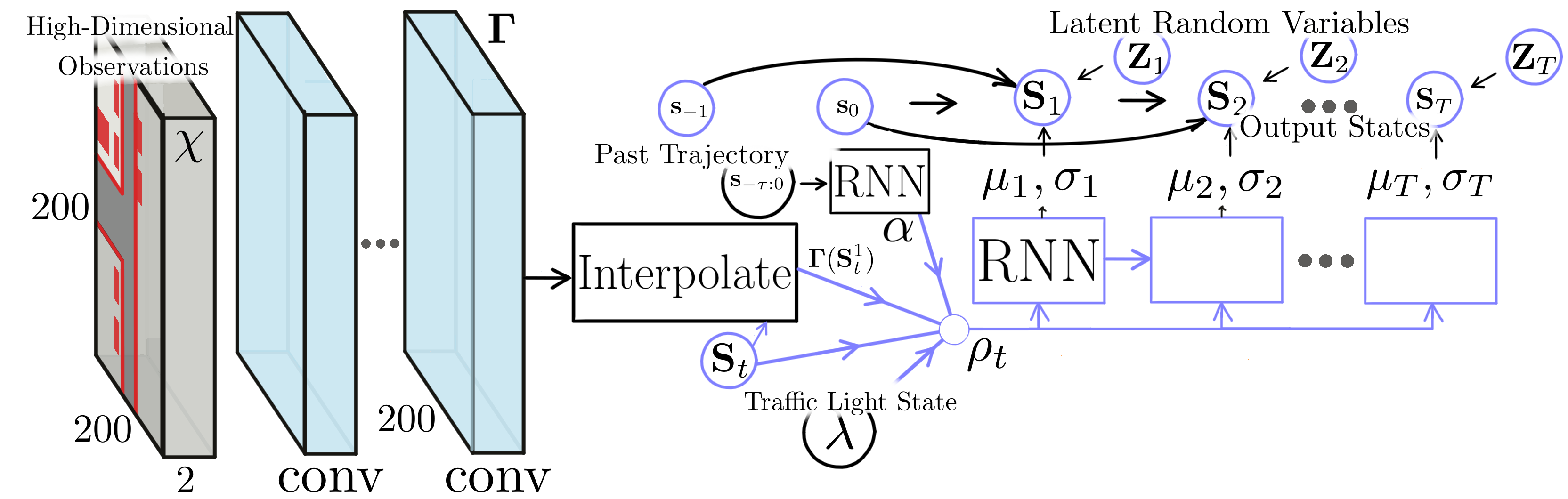}
\end{subfigure}
\caption{Architecture of $m_\theta$ and $\sigma_\theta$, which parameterize $q_\theta(\State|\phi\!=\!\{\chi,\state_{-\tau:0},\blambda\})$. Inputs: LIDAR $\chi$, past-states $\state_{-\tau:0}$, light-state $\blambda$, and latent noise $\Latent_{1:T}$. Output: trajectory $\State_{1:T}$. Details in Appendix~\ref{app:architecture}.} \label{fig:arch}
\end{figure}
In our autonomous driving application, we model the agent's state at time $t$ as $\state_t \in \mathbb{R}^{D}$ with $D\!=\!2$;  $\state_t$ represents our agent's location on the ground plane. The agent has access to environment perception $\phi \leftarrow \{\state_{-\tau:0}, \bchi, \blambda\}$, where $\tau$ is the number of past positions we condition on, $\bchi$ is a high-dimensional observation of the scene, and $\blambda$ is a low-dimensional traffic light signal. $\bchi$ could represent either LIDAR or camera images (or both), and is the agent's observation of the world. In our setting, we featurize LIDAR to $\bchi=\mathbb R^{200 \times 200 \times 2}$, with $\bchi_{ij}$ representing a 2-bin histogram of points above and at ground level in a $0.5\mathrm{m}^2$ cell at position $(i,j)$. CARLA  provides ground-truth $\state_{-\tau:0}$ and $\blambda$. Their availability is a realistic input assumption in perception-based autonomous driving pipelines. 

\noindent {\bf Model requirements:} A deep imitative model forecasts future expert behavior. It must be able to compute $q(\state\alltime|\phi)\forall \state\in\mathbb R^{T\times D}$. The ability to compute $\nabla_{\scriptsize \state\alltime}q(\state\alltime|\phi)$ enables gradient-based optimization for planning. \cite{rudenko2019human} provide a recent survey on forecasting agent behavior. As many forecasting methods cannot compute trajectory probabilities, we must be judicious in choosing $q(\State|\phi)$.
A model that can compute probabilities R2P2 \citep{rhinehart2018r2p2}, a generative autoregressive flow \citep{rezende2015variational,oord2017parallel}. 
We extend R2P2 to instantiate the deep imitative model $q(\State\alltime|\phi)$. {\bf R2P2  was previously used to forecast vehicle trajectories: it was not demonstrated or developed to  plan or execute controls.} Although we used R2P2, other future-trajectory density estimation techniques could be used -- designing $q(\state|\phi)$ is not the primary focus of this work. In R2P2, $q_\theta(\State\alltime|\phi)$ is induced by an invertible, differentiable function: $\State\!=\!f_\theta(\Latent;\phi)\!:\!\mathbb R^{T \times 2}\!\mapsto\!\mathbb R^{T \times 2}$; $f_\theta$ warps a latent sample from a base distribution $\Latent\!\sim\!q_0\!=\!\mathcal N(0, I)$ to $\State\alltime$. $\theta$ is trained to maximize $q_\theta(\State\alltime|\phi)$ of expert trajectories. $f_\theta$ is defined for  $1..T$ as follows:
\begin{align}
{ \State_t = f_t(\Latent_{1:t}) = \mu_\theta(\State_{1:t-1}, \phi) + \sigma_\theta(\State_{1:t-1}, \phi)\Latent_t},
\end{align} 
where $\mu_\theta(\State_{1:t-1}, \phi)\!=\!2\State_{t-1}\!-\!\State_{t-2}\!+\!m_\theta(\State_{1:t-1}, \phi)$ encodes a constant-velocity inductive bias. The $m_\theta \in \mathbb R^2$ and $\sigma_\theta \in \mathbb R^{2\times 2}$
are computed by expressive neural networks. The resulting trajectory distribution is complex and multimodal (Appendix~\ref{app:prior_visualization} depicts samples). Because traffic light state was not included in the $\phi$ of R2P2's ``RNN'' model, it could not react to traffic lights. We created a new model that includes $\blambda$. It fixed cases where $q(\State|\phi)$ exhibited no forward-moving preference when the agent was already stopped, and improved $q(\State|\phi)$'s stopping preference at red lights. We used $T\!=\!40$ trajectories at $10\mathrm{Hz}$ (4 seconds), and $\tau\!=\!3$. Fig.~\ref{fig:arch} depicts the architecture of $\mu_\theta$ and $\sigma_\theta$. 

\subsection{Imitative Driving}
\begin{figure*}[t]
    \centering
    \includegraphics[width=\textwidth]{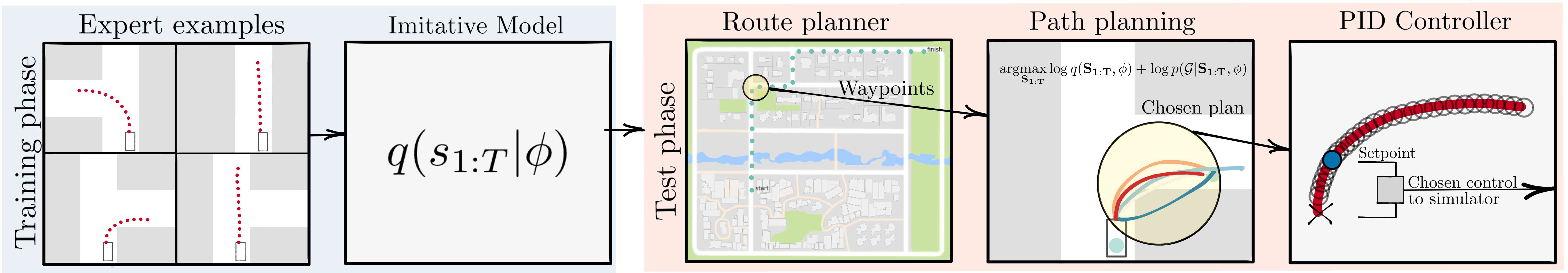}
    \caption{Illustration of our method applied to autonomous driving. Our method trains an imitative model from a dataset of expert examples. After training, the model is repurposed as an imitative planner. At test-time, a route planner provides waypoints to the imitative planner, which computes expert-like paths to each goal. The best plan is chosen according to the planning objective and provided to a low-level PID-controller in order to produce steering and throttle actions. This procedure is also described with pseudocode in Appendix~\ref{sec:pseudocode}.}\label{fig:application-pipeline}
\end{figure*}

We now instantiate a complete autonomous driving framework based on imitative models to study in our experiments, seen in Fig.~\ref{fig:application-pipeline}. We use three layers of spatial abstraction to plan to a faraway destination, common to autonomous vehicle setups: coarse route planning over a road map, path planning within the observable space, and feedback control to follow the planned path \citep{paden2016survey,schwarting2018planning}. For instance, a route planner based on a conventional GPS-based navigation system might output waypoints roughly in the lanes of the desired direction of travel, but not accounting for environmental factors such as the positions of other vehicles. This roughly communicates \emph{possibilities} of where the vehicle could go, but not \emph{when} or \emph{how} it could get to them, or any environmental factors like other vehicles. A goal likelihood from Sec.~\ref{sec:goallikelihoods} is formed from the route and passed to the planner, which generates a state-space plan according to the optimization in \Eqref{eqn:goal-sequence-planning}. 
The resulting plan is fed to a simple PID controller on steering, throttle, and braking. In 
Pseudocode of the driving, inference, and PID algorithms are given in Appendix~\ref{sec:pseudocode}. 

\section{Related Work}\label{sec:related_work}
A body of previous work has explored offline IL (Behavior Cloning -- BC) in the CARLA simulator \citep{li2018rethinking,liang2018cirl,sauer2018conditional,codevilla2018end,codevilla2019exploring}. These BC approaches 
condition on goals drawn from a small discrete set of directives. Despite BC's theoretical drift shortcomings \citep{ross2011dagger}, these methods still perform empirically well.
\emph{These approaches and ours share the same high-level routing algorithm: an A$^*$ planner on route nodes that generates waypoints}. In contrast to our approach, these approaches use the waypoints in a \emph{Waypoint Classifier}, which reasons about the map and the geometry of the route to classify the waypoints into one of several directives: \{Turn left, Turn right, Follow Lane, Go Straight\}. One of the original motivations for these type of controls was to enable \emph{a human} to direct the robot \citep{codevilla2018end}. {\bf However, in scenarios where there is no human in the loop (i.e. autonomous driving), we advocate for approaches to make use of the detailed spatial information inherent in these waypoints}. Our approach and several others we designed make use of this spatial information. One of these is CIL-States (CILS): whereas the approach in \cite{codevilla2018end} uses images to directly generate controls, \emph{CILS uses identical inputs and PID controllers as our method}. With respect to prior conditional IL methods, our main approach has more flexibility to handle more complex directives post-training, the ability to learn without goal labels, and the ability to generate  interpretable planned and unplanned trajectories. These contrasting capabilities are illustrated in Table~\ref{tab:baseline_attributes}. 

Our approach is also related to MBRL. MBRL can also plan, but with a one-step predictive model of \emph{possible} dynamics. The task of evoking expert-like behavior is offloaded to the reward function, which can be difficult and time-consuming to craft properly. We know of no MBRL approach previously applied to CARLA, so we devised one for comparison. \emph{This MBRL approach also uses identical inputs to our method}, instead to plan a reachability tree \citep{lavalle2006planning} over an dynamic obstacle-based reward function. See Appendix~\ref{app:baselines} for further details of the MBRL and CILS methods, which we emphasize use the \emph{same inputs} as our method. 

\begin{table}[t] 
\centering
    \begin{minipage}[b]{.99\textwidth}
    \centering
    \caption{Desirable attributes of each approach. A green check denotes that a method has a desirable attribute, whereas a red cross denotes the opposite.  A ``$^\dagger$'' indicates an approach we implemented.} \label{tab:baseline_attributes}
    \vspace{-3mm}
    \resizebox{\textwidth}{!}{
    \begin{tabular}{rccccccccc}
    \toprule
    Approach  & Flexible to New Goals & Trains without goal labels  & Outputs Plans  & Trains Offline & Has Expert P.D.F. \\
    \midrule
    {CIRL$^*$~\citep{liang2018cirl}   }  & \xmark & \xmark & \xmark & \xmark   & \xmark\\
    {CAL$^*$~\citep{sauer2018conditional} } & \xmark & \xmark & \xmark & \cmark & \xmark\\
    {MT$^*$~\citep{li2018rethinking}   }  & \xmark & \xmark & \xmark & \cmark  & \xmark \\
    {CIL$^*$~\citep{codevilla2018end} } & \xmark & \xmark & \xmark & \cmark   & \xmark\\
    {CILRS$^*$~\citep{codevilla2019exploring} }  & \xmark & \xmark & \xmark & \cmark  & \xmark\\
    {CILS$^\dagger$}  & \xmark & \cmark & \xmark & \cmark & \xmark\\
    {MBRL$^\dagger$} & \cmark & \cmark & \cmark & \xmark  & \xmark \\
    \rowcolor{ourmethod} {Imitative Models (\emph{Ours})$^\dagger$} & \cmark & \cmark & \cmark & \cmark & \cmark  \\
    \bottomrule
    \end{tabular}
    }
    \end{minipage}
    \vspace{1mm}
    \begin{minipage}[b]{.99\textwidth}
    \centering
    \caption{Algorithmic components of each approach.
    A ``$^\dagger$'' indicates an approach we implemented. 
    } \label{tab:baseline_components}
    \vspace{-3mm}
    \resizebox{\textwidth}{!}{
    \begin{tabular}{rl@{\hskip 6pt}c@{\hskip 6pt}l@{\hskip 6pt}c@{\hskip 6pt}l@{\hskip 6pt}c@{\hskip 6pt}ll}
    \toprule
    Approach     & Control Algorithm &  $\scriptsize\leftarrow\!$ & Learning Algorithm & $\leftarrow$ &  Goal-Generation Algorithm &  $\leftarrow$ & Routing Algorithm & High-Dim. Obs. \\  
    \midrule
    {CIRL$^*$~\citep{liang2018cirl}   }  & Policy & &   Behavior Cloning+RL  & & Waypoint Classifier & & A$^*$ Waypointer  & Image\\ 
    {CAL$^*$~\citep{sauer2018conditional} }  & PID & & Affordance Learning &  & Waypoint Classifier & & A$^*$ Waypointer & Image\\
    {MT$^*$~\citep{li2018rethinking}   }  & Policy & &  Behavior Cloning & & Waypoint Classifier & & A$^*$ Waypointer & Image\\
    {CIL$^*$~\citep{codevilla2018end} } & Policy & &  Behavior Cloning & & Waypoint Classifier & & A$^*$ Waypointer & Image\\ 
    {CILRS$^*$~\citep{codevilla2019exploring} } &  Policy & &  Behavior Cloning & & Waypoint Classifier & & A$^*$ Waypointer & Image\\
    {CILS$^\dagger$ } & PID & & Trajectory Regressor &  & Waypoint Classifier & & A$^*$ Waypointer& LIDAR  \\
    {MBRL$^\dagger$}  &  Reachability Tree & & State Regressor & & Waypoint Selector & & A$^*$ Waypointer& LIDAR \\
    \rowcolor{ourmethod} {Imitative Models (\emph{Ours})$^\dagger$} & Imitative Plan+PID & & Traj. Density Est.  & & Goal Likelihoods  & & A$^*$ Waypointer& LIDAR\\
    \bottomrule
    \vspace{2mm} 
    \end{tabular} }
    \end{minipage} 
    \vspace{-9mm}  
\end{table}

Several prior works~\citep{tamar2016value,Amos-NIPS-18,srinivas2018universal} used imitation learning to train policies that contain planning-like modules as part of the model architecture. While our work also combines planning and imitation learning, ours captures a distribution over possible trajectories, and then plan trajectories at test-time that accomplish a variety of given goals with high probability under this distribution. Our approach is suited to offline-learning settings where interactively collecting data is costly (time-consuming or dangerous). However, there exists online IL approaches that seek to be \emph{safe} \citep{menda2017dropoutdagger,sun2017fast,zhang2017query}. 

\section{Experiments} \label{sec:experiments}
We evaluate our method using the CARLA driving simulator \citep{dosovitskiy17carla}. We seek to answer four primary questions: {\bf (1) Can we generate interpretable, expert-like plans with offline learning and no reward engineering}? Neither IL nor MBRL can do so. It is straightforward to \emph{interpret} the trajectories by visualizing them on the ground plane; we thus seek to validate whether these plans are \emph{expert-like} by equating expert-like behavior with high performance on the CARLA benchmark. {\bf (2) Can we achieve state-of-the-art CARLA performance using resources commonly available in real autonomous vehicle settings?} \emph{There are several differences between the approaches, as discussed in Sec~\ref{sec:related_work} and shown in Tables~\ref{tab:baseline_attributes}~and~\ref{tab:baseline_components}. Our approach uses the CARLA toolkit's resources that are commonly available in real autonomous vehicle settings: waypoint-based routes (all prior approaches use these) and LIDAR  (CARLA-provided, but only the approaches we implemented use it). Furthermore, the two additional methods of comparison we implemented (CILS and MBRL) use the exact same inputs as our algorithm. These reasons justify an overall performance comparison to answer (2): whether we can achieve state-of-the-art performance using commonly available resources. We advocate that other approaches also make use of such resources.}
{\bf (3) How flexible is our approach to new tasks?} We investigate (3) by applying each of the goal likelihoods we derived and observing the resulting performance. {\bf (4) How robust is our approach to error in the provided goals?} We do so by injecting two different types of error into the waypoints and observing the resulting performance.
 

We begin by training $q(\State|\phi)$ on a dataset of $25$ hours of driving we collected in \texttt{Town01}, detailed in Appendix~\ref{app:dataset}. Following existing protocol, each test episode begins with the vehicle randomly positioned on a road in the \texttt{Town01} or \texttt{Town02} maps in one of two settings: static-world (no other vehicles) or dynamic-world (with other vehicles). 
We construct the goal set $\mathbb G$ for the Final-State Indicator (A) directly from the route output by CARLA's waypointer. B's line segments are formed by connecting the waypoints to form a piecewise linear set of segments. C's regions are created a \emph{polygonal goal region} around the segments of (B). Each represents an increasing level of coarseness of direction. Coarser directions are easier to specify when there is ambiguity in positions (both the position of the vehicle and the position of the goals). Further details are discussed in Appendix~\ref{app:goalsetconstruction}.
We use three metrics: 
(a) success rate in driving to the destination without any collisions (which all prior work reports); 
(b) red-light violations; and
(c) proportion of time spent driving in the wrong lane and off road.
With the exception of metric (a), lower numbers are better. 

\noindent{\bf Results}: Towards questions (1) and (3) (expert-like plans and flexibility), we apply our approach with a variety of goal likelihoods to the CARLA simulator. Towards question (2), we compare our methods against CILS, MBRL, and prior work. These results are shown in Table~\ref{table:dynmetrics}. The metrics for the methods we did not implement are from the aggregation reported in \cite{codevilla2019exploring}. We observe our method to outperform all other approaches in all settings: static world, dynamic world, training conditions, and test conditions. We observe the \emph{Goal Indicator methods are able to perform well, despite having no hyperparameters to tune}. We found that we could further improve our approach's performance if we use the light state to define different goal sets, which defines a ``smart'' waypointer. The settings where we use this are suffixed with ``S.'' in the Tables. We observed the planner prefers \emph{closer} goals when obstructed, when the vehicle was already stopped, and when a red light was detected; we observed the planner prefers \emph{farther} goals when unobstructed and when green lights or no lights were observed. Examples of these and other interesting behaviors are best seen in the videos on the website ({\scriptsize \url{https://sites.google.com/view/imitative-models}}). These behaviors follow from the method leveraging $q(\State|\phi)$'s internalization of aspects of expert behavior in order to reproduce them in new situations. Altogether, these results provide affirmative answers to questions (1) and (2). Towards question (3), these results show that our approach is flexible to different directions defined by these goal likelihoods. 

\newcommand\hfilll{}

\begin{table*}[htb]
\centering
\caption{We evaluate different autonomous driving methods on CARLA's \emph{Dynamic Navigation} task. A ``$^\dagger$'' indicates methods we have implemented (each observes the same waypoints and LIDAR as input). A ``$^*$'' indicates results reported in \cite{codevilla2019exploring}. A ``--'' indicates an unreported statistic. A ``$^\ddagger$'' indicates an optimistic estimate in transferring a result from the static setting to the dynamic setting. ``S.'' denotes a ``smart'' waypointer reactive to light state, detailed in Appendix~\ref{app:smartwaypointer}.}
\label{table:dynmetrics}
\resizebox{\textwidth}{!}{
 \begin{tabular}{l                         c    c     D               D                 c         c        D       D    D       D        D l  c      c   D               D                 c         c        D       D     D       D        D} 
 \toprule
  & \multicolumn{10}{c}{\texttt{Town01} (training conditions)} &  \multicolumn{10}{c}{\texttt{Town02} (test conditions)}  \vspace{1mm}  \\
\cline{2-9}  \cline{14-21} \vspace{-3mm}\\ 
 Dynamic Nav. Method & Success & Ran Red Light & Success & Collision Impulse &  Wrong lane & Off road & Accel & Jerk & Snap & Crackle & Pop & & Success & Ran Red Light & Success & Collision Impulse & Wrong lane & Off road & Accel & Jerk & Snap & Crackle & Pop \\
 \cline{1-9}  \cline{14-21} \vspace{-2mm}\\ 
\rowcolor{priormethod}   CIRL$^*$\hfilll \citep{liang2018cirl} & 82\% & --& -- & -- & -- & -- & -- &-- &-- &--& & &  41\%& -- & --  & -- & --& -- & -- & --  \\
 \rowcolor{priormethod}    CAL$^*$\hfilll \citep{sauer2018conditional} & 83\% & --& -- & -- & -- & -- & -- &-- &-- &--& & & 64\% & -- & --  & -- & --& -- & -- & --  \\
  \rowcolor{priormethod}   MT$^*$ \hfilll\citep{li2018rethinking} & 81\% & -- & -- & -- & -- & -- & -- &-- &-- &--& & &53\% & --& --  & -- & --& -- & -- & --  \\
  \rowcolor{priormethod}        CIL$^*$\hfilll \citep{codevilla2018end} & 83\% & 83\%$^\ddagger$& -- & -- & -- & -- & -- &-- &-- &--& & &   38\% & 82\%$^\ddagger$ & --  & -- & --& -- & -- & --  \\
 \rowcolor{priormethod}    CILRS$^*$\hfilll\citep{codevilla2019exploring} & 92\%&  27\%$^\ddagger$   & -- & -- & -- & -- & -- &-- &-- &--& & & {66\%} & 64\%$^\ddagger$   &-- & --  & -- & --& -- & -- & --  \\
  CILS, Waypoint Input$^\dagger$  & 17\%  & {\bf 0.0\%} &  &  & 0.20\% & 12.1\% &FF & 23.0 &--& & &   & 36\% & {\bf 0.0}\% & \xmark & \xmark & 1.11\% & 11.70\% & \xmark & 44.2 & \xmark & \xmark & \xmark \\
   MBRL, Waypoint Input$^\dagger$ & 64\% & 72\% & \xmark & \xmark & 11.1\% & 2.96\% & \xmark & 39.9 & \xmark & \xmark & \xmark & & 48\% & 54\% & \xmark & \xmark & 20.6\% & 13.3 \% & \xmark & 44.2 & \xmark & \xmark & \xmark \\
   \rowcolor{ourmethod} \textit{Our method, Final-State Indicator}$^\dagger$ & 92\% & 26\% & 92\% & 1.394 & 0.05\% & 0.012\% & 4.122 & 18.945 & 276.23 & 4727.7 & 92384& & 84\% & 35\% & 84\% & 709.128 & 0.13\% & 0.38\% & 4.583 & 29.214 & 489.14 & 9141.5 & 175619 \\
   \rowcolor{ourmethod} \textit{Our method, Line Segment Final-St. Indicator}$^\dagger$ & 84\% & 42\% & 84\% & 1.732 & {\bf 0.03\%} & 0.295\% & 3.978 & 18.880 & 251.54 & 4515.2 & 80740 & & 88\% & 33\% & 88\% & 4923.844 & 0.12\% & 0.14\% & 4.260 & 23.015 & 327.75 & 5703.2 & 107852 \\
   \rowcolor{ourmethod} \textit{Our method, Region Final-St. Indicator}$^\dagger$ & 84\% & 56\% & 84\% & 513.947 & {\bf 0.03\%} & 0.131\% & 3.331 & 15.648 & 224.21 & 4023.1 & 73648 && 88\% & 54\% & 88\% & 1135.528 & 0.13\% & 0.22\% & 3.679 & 18.190 & 278.53 & 4741.3 & 85966 \\
        \rowcolor{ourmethod} \textit{Our method, Gaussian Final-St. Mix.}$^\dagger$  & {92\%} & {6.3\%} & {\bf 25 / 25} & 5.698 & {0.04\%} & {\bf 0.005\%} & & 24.1 & & a & b& & {\bf 100}\%& {12\%} & &  &{ 0.11\%} &{\bf 0.04\%} & X & \xmark \\
   \rowcolor{ourmethod} \textit{Our method, Region Final-St. Indicator  S.}$^\dagger$ & 92\% & 2.8\% & 92\% & 255.032 & 0.021\% & 0.099\% & 4.307 & 19.408 & 285.00 & 5000.5 & 98412 && 92\% & 4.0\% & 92\% & 24969.125 & 0.11\% & 1.85\% & 4.564 & 22.558 & 330.25 & 5989.9 & 113906 \\
  \rowcolor{ourmethod} \textit{Our method, Gaussian Final-St. Mix. S.}$^\dagger$  & {\bf 100\%} & 1.7\% & {\bf 25 / 25} & 5.698 & {\bf 0.03\%} & {\bf 0.005\%} & & \xmark & & a & b& & { 92}\%& {\bf 0.0\%} & &  &{\bf 0.05\%} &{ 0.15\%} & X & 30.2\\
 \bottomrule
 \end{tabular} 
 }
 \resizebox{\textwidth}{!}{
 \begin{tabular}{l                         c    c     D               D                 c         c        D       D    D       D        D l  c      c   D               D                 c         c        D       D     D       D        D} 
  & \multicolumn{10}{c}{\texttt{Town01} (training conditions)} &  \multicolumn{10}{c}{\texttt{Town02} (test conditions)}  \vspace{1mm}  \\
\cline{2-9}  \cline{14-21} \vspace{-3mm}\\ 
 Static Nav. Method & Success & Ran Red Light & Success & Collision Impulse &  Wrong lane & Off road & Accel & Jerk & Snap & Crackle & Pop & & Success & Ran Red Light & Success & Collision Impulse & Wrong lane & Off road & Accel & Jerk & Snap & Crackle & Pop \\
  \cline{1-9}  \cline{14-21} \vspace{-2mm}\\ 
     \rowcolor{priormethod} CIRL$^*$ \hfilll\citep{liang2018cirl} & 93\% & -- & -- & -- & -- & -- & -- & -- & & & & &  68\% & --  & -- & -- & --  & --& --& -- \\
\rowcolor{priormethod}    CAL$^*$ \hfilll\citep{sauer2018conditional} & 92\% & -- & -- & -- & -- & -- & -- & -- & & & & & 68\% &-- &-- &--& --& --& -- & -- \\
 \rowcolor{priormethod}   MT$^*$ \hfilll\citep{li2018rethinking} & 81\%  & -- & -- & -- & -- & -- & -- & -- & & & & & 78\% & -- & -- & -- & -- & -- & -- & -- \\
  \rowcolor{priormethod}   CIL$^*$ \hfilll\citep{codevilla2018end} & 86\% & 83\% & n / n  & -- & -- & --  & -- & -- & -- & & & & 44\% & 82\% & --  & --& --& --& -- & --\\
 \rowcolor{priormethod}   CILRS$^*$\hfilll\citep{codevilla2019exploring} & 95\% & 27\% &  -- & -- & -- & -- & -- & -- & & & && 90\% & 64\% & -- & -- & -- & -- & -- & -- \\
  CILS, Waypoint Input$^\dagger$  & 28\% & {\bf 0.0}\% & 7 / 10 & 0.96 & 0.38\% & 10.23\%  & 0.203 & \xmark & 5.52 & 46.9 & 438 & & 36\%& {\bf 0.0}\% & ZZ & 3 / 10 & 1.69\% & 16.82\% & 3.06\%  & 17.1 & 1.234 & 8.13 & 91.1  \\
  MBRL, Waypoint Input$^\dagger$ & { 96\%}& 78\% & { 10 / 10} & { 0.0} & 14.3\% & 1.94\% & 0.062 & 39.7 & 2.69 & 26.1 & 261 & & { 96\%} & 73\% & \xmark & \xmark  & 19.6 \% & 0.75\%  & \xmark & 42.7 & \xmark & \xmark & \xmark \\ 
  \rowcolor{ourmethod} \textit{Our method, Final-State Indicator}$^\dagger$ & {\bf 100\%} & 48\% & 100\% & 1.083 & 0.05\% & {\bf 0.002\%} & 3.717 & 22.340 & 378.53 & 6924.6 & 140192 & & {\bf 100\%} & 52\% & 100\% & 1415.731 & 0.10\% & {\bf 0.13\%} & 4.144 & 26.482 & 466.42 & 8406.1 & 169602 \\
    \rowcolor{ourmethod}
  \textit{Our method, Gaussian Final-St. Mixture}$^\dagger$                 & { 96}\% &  {0.83}\% & { 24 / 25} & {\bf 0.0} & {\bf 0.01}\% & 0.08\% & {\bf 0.054}  & {24.0} &  {\bf 1.50}  &  {\bf 13.8}  & {\bf 136}    &  & {96\%} & {\bf 0.0\%} & {\bf 8 / 10} & {\bf 0.41} & {\bf 0.03\%} &  {0.14\%} & {\bf  0.054 } & {31.0} & {\bf 2.64} & {\bf 21.4} & {\bf 289} \\ 
  \rowcolor{ourmethod}
    \textit{Our method, Gaussian Final-St. Mix. S.}$^\dagger$                 & { 96}\% & {\bf 0.0\%} & {\bf 24 / 25} & {\bf } &  {0.04\%} & { 0.07\%} & {\bf 0.054}  & \xmark &  {\bf 1.50}  &  {\bf 13.8}  & {\bf 136}    &  & {92\%} & {\bf 0.0\%} & {\bf 8 / 10} & {\bf 00} & {0.18\%} &  {  0.27\%} & {\bf  0.054 } & 22.7 & {\bf DD} & {\bf 21.4} & {\bf 289} \\ 
    \bottomrule
    \end{tabular} }
\end{table*}

\subsection{Robustness to Errors in Goal-Specification} \label{sec:noise_robustness}

Towards questions (3) (flexibility) and (4) (noise-robustness), we analyze the performance of our method when the path planner is heavily degraded, to understand its stability and reliability. We use the \emph{Gaussian Final-State Mixture} goal likelihood.

\newlength{\roblength}
\setlength{\roblength}{.185\columnwidth}
\begin{figure}[tbh]
\begin{minipage}[t]{.79\textwidth}
    \centering
    \begin{subfigure}[t]{\roblength}
       \begin{overpic}[width=\textwidth]{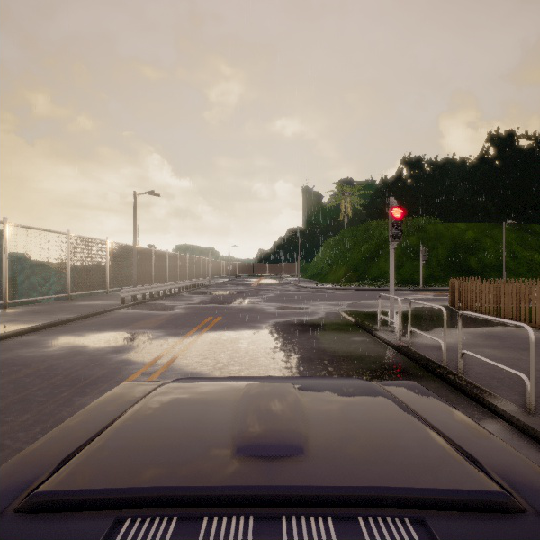} 
       \end{overpic}
    \end{subfigure}
    \begin{subfigure}[t]{\roblength}
       \FBox{ 
       \begin{overpic}[width=\textwidth,angle=90]{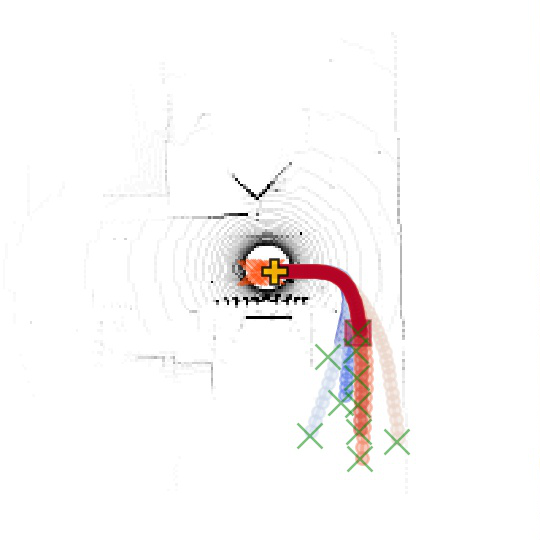} 
       \put(0,15){\includegraphics[width=0.7\textwidth]{img/legends/teaser_plotlegend.png}}
      \put(0,0){\includegraphics[width=0.7\textwidth]{img/legends/coolwarm_lth.png}}
       \end{overpic}
       }
    \end{subfigure}
          \begin{subfigure}[t]{\roblength}
       \begin{overpic}[width=\textwidth]{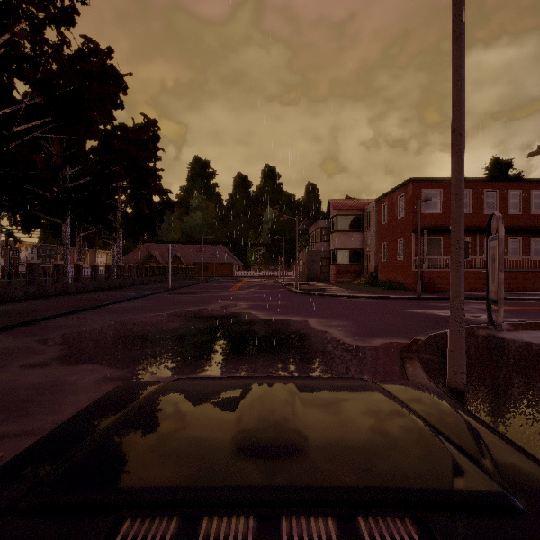}
       \end{overpic}
    \end{subfigure}
    \FBox{ 
      \begin{subfigure}[t]{\roblength}
       \begin{overpic}[width=\textwidth,angle=90]{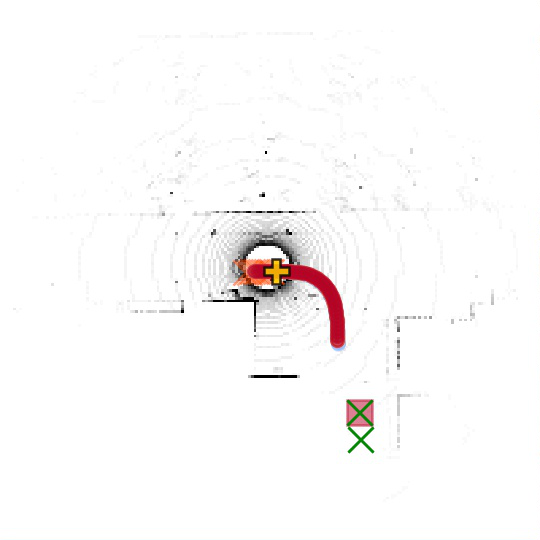}
       \put(0,15){\includegraphics[width=0.7\textwidth]{img/legends/teaser_plotlegend.png}}
      \put(0,0){\includegraphics[width=0.7\textwidth]{img/legends/coolwarm_lth.png}}
       \end{overpic}
    \end{subfigure} 
    }
       
    \caption{Tolerating bad goals. The planner prefers goals in the distribution of expert behavior (on the road at a reasonable distance). \emph{Left}: Planning with $\nicefrac{1}{2}$ decoy goals. \emph{Right}: Planning with all goals on the wrong side of the road.} \label{fig:decoy-waypoints}
\end{minipage}
\hfill
\begin{minipage}[t]{\roblength}
    \centering
             \FBox{
     \begin{overpic}[width=\textwidth,angle=90,trim={5cm 6cm 2cm 0cm},clip]{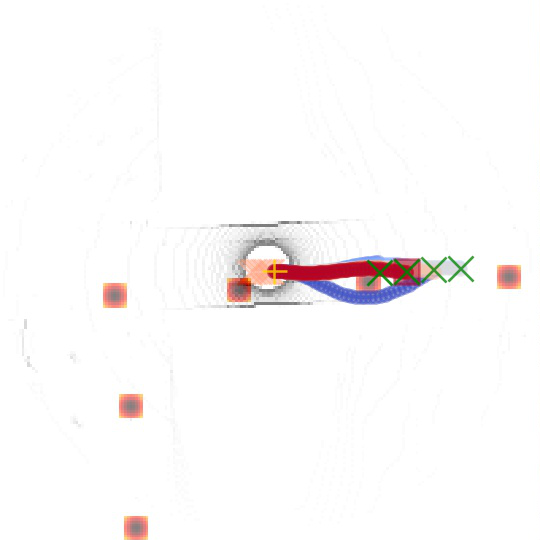}
         \put(25,40){\includegraphics[width=0.3\textwidth]{img/legends/teaser_plotlegend.png}}
      \put(18,20){\includegraphics[width=0.42\textwidth]{img/legends/coolwarm_lth.png}}
     \end{overpic}
     }
    \caption{Test-time plans steering around potholes.}
    \label{fig:sparse-potholes}
\end{minipage}  
\end{figure}

\noindent{\bf Navigating with high-variance waypoints.} As a test of our model's capability to stay in the distribution of demonstrated behavior, we designed a ``decoy waypoints'' experiment, in which \emph{half} of the waypoints are highly perturbed versions of the other half, serving as distractions for our Gaussian Final-State Mixture imitative planner. We observed surprising robustness to decoy waypoints. Examples of this robustness are shown in Fig.~\ref{fig:decoy-waypoints}. In Table~\ref{tab:decoy-waypoints}, we report the success rate and the mean number of planning rounds for failed episodes in the ``\nicefrac{1}{2} distractors'' row. These numbers indicate our method can execute dozens of planning rounds without decoy waypoints causing a catastrophic failure, and often it can execute the hundreds necessary to achieve the goal. See Appendix~\ref{app:robustness} for details. 

\noindent{\bf Navigating with waypoints on the wrong side of the road.} We also designed an experiment to test our method under systemic bias in the route planner. Our method is provided waypoints on the wrong side of the road (in CARLA, the left side), and tasked with following the directions of these waypoints while staying on the \emph{correct} side of the road (the right side). In order for the value of $q(\state|\phi)$ to outweigh the influence of these waypoints, we increased the $\epsilon$ hyperparameter.
We found our method to still be very effective at navigating, and report results in Table~\ref{tab:decoy-waypoints}. We also investigated providing very coarse 8-meter wide regions to the Region Final-State likelihood; these always include space in the wrong lane and off-road (Fig.~\ref{fig:bigregionvis} in Appendix~\ref{app:goallikelihoodvis} provides visualization). Nonetheless, on Town01 Dynamic, this approach still achieved an overall success rate of $48\%$.  Taken together towards question (4), our results indicate that our method is \emph{fairly robust to errors in goal-specification}.

 
\subsection{Producing Unobserved Behaviors to Avoid Novel Obstacles} \label{sec:pothole_experiments}

\newlength{\potholelength}
\setlength{\potholelength}{.192\columnwidth}
\begin{table}[tbh]
\centering
\caption{\textbf{Robustness to waypoint noise and test-time pothole adaptation}. Our method is robust to waypoints on the wrong side of the road  and fairly robust to decoy waypoints. Our method is flexible enough to safely produce behavior not demonstrated (pothole avoidance) by incorporating a test-time cost. Ten episodes are collected in each Town.}
\label{tab:decoy-waypoints}
\resizebox{.8\textwidth}{!}
{
\begin{tabular}{rcccDDDcDccDDDc}
\toprule
  & & \multicolumn{6}{c}{\texttt{Town01} (training conditions)} &  &  \multicolumn{6}{c}{\texttt{Town02} (test conditions)}  \vspace{1mm} \\
 \cline{3-8}  \cline{10-15}
  Waypointer & Extra Cost & Success  & Wrong lane & Off road &  \#plans until success & \#plans until failure & Potholes hit & & Success  & Wrong lane & Off road &  \#plans until success & \#plans until failure & Potholes hit \\
  \midrule
  Noiseless waypointer &  & {100\%} & 0.00\% & 0.0\%  & & &  177/230 &  & {100\%} & 0.41\% & 0.4\% & & & 82/154\\
  Waypoints wrong lane  &  & {100\%} & 0.34\% & 0.0\%  & {--} & {--}& {--}&  & {70\%} & 3.16\% & 0.0\% & --  & -- & -- \\
  $\nicefrac{1}{2}$ waypoints distracting &  & {70\%} & {--} & {--} & 157.6 & 37.9& & {--} & {50\%} & {--} & {--} & 78.0 & 32.1& --\\
    Noiseless waypointer & Pothole & {90\%} & 1.53\% & 0.1\% &  & &{10/230}  &  & {70\%} & 1.53\% & { 0.1\%} &  & &{35/154} \\
  \bottomrule
\end{tabular}
}
\end{table}

To further investigate our model's flexibility to test-time objectives (question 3), we designed a pothole avoidance experiment. We simulated potholes in the environment by randomly inserting them in the cost map near waypoints.
We ran our method with a test-time-only cost map of the simulated potholes by combining goal likelihoods (F) and (G), and compared to our method that did not incorporate the cost map (using (F) only, and thus had no incentive to avoid potholes). We recorded the number of collisions with potholes. In Table~\ref{tab:decoy-waypoints}, our method with cost incorporated avoided most potholes while avoiding collisions with the environment. To do so, it drove closer to the centerline, and occasionally entered the opposite lane. Our model internalized obstacle avoidance by staying on the road and demonstrated its flexibility to obstacles not observed during training. Fig.~\ref{fig:sparse-potholes} shows an example of this behavior. See Appendix~\ref{sec:potholedetails} for details of the pothole generation.

\section{Discussion}

We proposed ``Imitative Models'' to combine the benefits of IL and MBRL. Imitative Models are probabilistic predictive models able to plan interpretable expert-like trajectories to achieve new goals. Inference with an Imitative Model resembles trajectory optimization in MBRL, enabling it to both \emph{incorporate new goals} and \emph{plan to them} at test-time, which IL cannot. Learning an Imitative Model resembles offline IL, enabling it to circumvent the difficult reward-engineering and costly online data collection necessities of MBRL. We derived families of flexible goal objectives and showed our model can successfully incorporate them without additional training. Our method substantially outperformed six IL approaches and an MBRL approach in a dynamic simulated autonomous driving task.  We showed our approach is robust to poorly specified goals, such as goals on the wrong side of the road. We believe our method is broadly applicable in settings where expert demonstrations are available, flexibility to new situations is demanded, and safety is paramount. 
\bibliography{bibliography}
\bibliographystyle{iclr2020_conference}

\clearpage
\appendix
 
\section{Algorithms} \label{sec:pseudocode}

\begin{algorithm}
    \caption{\, \sc{ImitativeDriving}({\fontsize{8.5}{8.5}\textsc{RoutePlan}, \textsc{ImitativePlan}, \textsc{PIDController}}, $q_\theta, H$)}
    \label{alg:control}
    \begin{algorithmic}[1]
    \STATE $\phi \gets \textsc{Environment}(\emptyset)$ \COMMENT{Initialize the robot}
    \WHILE{not at destination}
    \STATE $\mathcal G \gets \textsc{RoutePlan}(\phi)$ \COMMENT{Generate goals from a route}
    \STATE $\state^{\mathcal G}_{1:T} \gets \textsc{ImitativePlanR2P2}(q_\theta, \mathcal G, \phi)$ \COMMENT{Plan path}
    \FOR{$h=0$ to $H$} 
    \STATE    $u \gets \textsc{PIDController}(\phi, \state^{\mathcal G}_{1:T}, h, H)$ 
    \STATE    $\phi \gets \textsc{Environment}(u)$ \COMMENT{Execute control}
    \ENDFOR
    \ENDWHILE
  \end{algorithmic}
\end{algorithm}

In Algorithm~\ref{alg:control}, we provide pseudocode for receding-horizon control via our imitative model. In Algorithm~\ref{alg:latent-planning} we provide pesudocode that describes how we plan in the latent space of the trajectory. Since $\state_{1:T} = f(\latent_{1:T})$ in our implementation, and $f$ is differentiable, we can perform gradient descent of the same objective in terms of $\latent_{1:T}$. Since $q$ is trained with $\latent_{1:T}\sim \mathcal N(0, I)$, the latent space is likelier to be better numerically conditioned than the space of $\state_{1:T}$, although we did not compare the two approaches formally. 
\begin{algorithm}
    \caption{\, \sc{ImitativePlanR2P2}($q_\theta, \mathcal G,  \phi, f$)}
    \label{alg:latent-planning}
    \begin{algorithmic}[1]
        \STATE Define MAP objective $\mathcal L$ with $q_\theta$ according to \Eqref{eqn:goal-sequence-planning} \COMMENT{Incorporate the Imitative Model}
        \STATE Initialize $\latent_{1:T} \sim q_0$
        \WHILE{not converged} 
            \STATE $\latent_{1:T} \gets \latent_{1:T} + \nabla_{\latent_{1:T}}\mathcal L(\state_{1:T}=f(\latent_{1:T}), \mathcal G,  \phi) $
        \ENDWHILE
        \STATE \textbf{return} $\state_{1:T}=f(\latent_{1:T})$
    \end{algorithmic}
\end{algorithm}

In Algorithm~\ref{alg:pid-controller}, we detail the speed-based throttle and position-based steering PID controllers.

\begin{algorithm}
    \caption{\, \sc{PIDController}($\phi=\{\state_{0}, \state_{-1},\dots\}, \state^{\mathcal G}_{1:T}, h, H; K_p^{\dot{s}}, K_p^{\alpha}$)}
    \label{alg:pid-controller}
    \begin{algorithmic}[1]
    \STATE $i\leftarrow T - H + h$ \COMMENT{Compute the index of the target position}
    \STATE $\dot{s}_{\text{process-speed}}\leftarrow(\state_{0,x}-\state_{-1,x})$ \COMMENT{Compute the current forward speed from the observations}
        \STATE $s_{\text{setpoint-position}}\leftarrow \state^{\mathcal G}_{i,x}$ \COMMENT{Retrieve the target position x-coordinate from the plan}
        \STATE $\dot{s}_{\text{setpoint-speed}}\leftarrow\nicefrac{s_{\text{setpoint-position}}}{i}$ \COMMENT{Compute the forward target speed}
        \STATE $e_{\dot{s}}\leftarrow\dot{s}_{\text{setpoint-speed}}-\dot{s}_{\text{process-speed}}$ \COMMENT{Compute the forward speed error}
        \STATE $u_{\dot{s}}\leftarrow K_p^{\dot{s}} e_{\dot{s}}$ \COMMENT{Compute the accelerator control with a nonzero proportional term}
        \STATE throttle $\leftarrow\mathbbm 1(e > 0) \cdot u + \mathbbm 1(e \leq 0) \cdot 0$ \COMMENT{Use the control as throttle if the speed error is positive}
        \STATE brake $\leftarrow\mathbbm 1(e > 0) \cdot 0 + \mathbbm 1(e \leq 0) \cdot u$ \COMMENT{Use the control as brake if the speed error is negative}
        \STATE $\alpha_{\text{process}}\leftarrow\arctan(\state_{0,y}-\state_{-1,y},\state_{0,x}-\state_{-1,x})$ \COMMENT{Compute current heading}
        \STATE $\alpha_{\text{setpoint}}\leftarrow\arctan(\state^{\mathcal G}_{i,y}-\state_{0,y},|\state^{\mathcal G}_{i,x}-\state_{0,x}|)$ \COMMENT{Compute target forward heading}
        \STATE $e_\alpha\leftarrow\alpha_{\text{setpoint}}-\alpha_{\text{process}}$ \COMMENT{Compute the heading error}
        \STATE steering $\leftarrow K_p^\alpha e_\alpha$ \COMMENT{Compute the steering with a nonzero proportional term}
        \STATE $u\leftarrow\left[\text{throttle}, \text{steering}, \text{brake}\right]$
        \STATE \textbf{return} $u$
    \end{algorithmic}
\end{algorithm}

\section{Goal Details} \label{app:goaldetails}

\subsection{Optimizing Goal Likelihoods with Set Constraints} \label{app:goalsetlikelihoods}
We now derive an approach to optimize our main objective with set constraints. Although we could apply a constrained optimizer, we find that we are able to exploit properties of the model and constraints to derive differentiable objectives that enable approximate optimization of the corresponding closed-form optimization problems. These enable us to use the same straightforward gradient-descent-based optimization approach described in Algorithm~\ref{alg:latent-planning}.

\paragraph{Shorthand notation:} In this section we omit dependencies on $\phi$ for brevity, and use
short hand $\mu_t\doteq\mu_\theta(\state_{1:t-1})$ and $\Sigma_t\doteq\Sigma_\theta(\state_{1:t-1})$. For example, 
$q(\state_t|\state_{1:t-1})=\mathcal N\left(\state_t;\mu_t,\Sigma_t\right)$. 

Let us begin by defining a useful delta function: 
\begin{align}
    \delta_{\state_T}(\mathbb G) 
    \;\doteq\; \begin{cases} 1 & \text{if}\;\; \state_T \in \mathbb G \\ 0 & \text{if}\;\; \state_T \not\in \mathbb G, \end{cases}
\end{align}
which serves as our goal likelihood when using goal with set constraints: $p(\mathcal G | \state_{1:T})\leftarrow\delta_{S_T}(\mathbb G)$. We now derive the corresponding maximum a posteriori optimization problem:
\begin{align}
    \state_{1:T}^*
    &\;\doteq\; \argmax_{\state_{1:T}\in \mathbb R^{2T}}\; p(\state_{1:T}|\mathcal G)  \nonumber \\
    &\;=\; \argmax_{\state_{1:T}\in \mathbb R^{2T}}\; p(\mathcal G | \state_{1:T}) \cdot q(\state_{1:T}) \cdot p^{-1}(\mathcal G) \nonumber \\ 
    &\;=\; \argmax_{\state_{1:T}\in \mathbb R^{2T}}\; \underbrace{p(\mathcal G | \state_{1:T})}_{\text{goal likelihood}} \cdot \underbrace{q(\state_{1:T})}_{\text{imitation prior}} \nonumber \\
    &\;=\; \argmax_{\state_{1:T}\in \mathbb R^{2T}}\; \underbrace{\delta_{S_T}(\mathbb G)}_{\text{set constraint}} \cdot \underbrace{q(\state_{1:T})}_{\text{imitation prior}} \nonumber \\ 
    &\;=\; \argmax_{\state_{1:T}\in \mathbb R^{2T}}\; \begin{cases}q(\state_{1:T}) & \text{if}~ \state_T \in \mathbb G\\
    0 &\text{if}~ \state_T \not\in \mathbb G
    \end{cases} \nonumber \\
    &\;=\; \argmax_{\state_{1:T-1}\in \mathbb R^{2(T-1)}, \state_T \in \mathbb G} q(\state_{1:T}) \nonumber \\
    &\;=\; \argmax_{\state_{1:T-1}\in \mathbb R^{2(T-1)}} \argmax_{\state_T \in \mathcal G}\;\;  q(\state_T|\state_{1:T-1}) \prod_{t=1}^{T-1} q(\state_t|\state_{1:t-1}) \nonumber \\ 
    &\;=\; \argmax_{\state_{1:T-1}\in \mathbb R^{2(T-1)}} \argmax_{\state_T \in \mathcal G}\;\; \mathcal N(\state_T;\mu_T,\Sigma_T) \prod_{t=1}^{T-1} \mathcal N(\state_t;\mu_t,\Sigma_t). \label{eq:sAllstar}
\end{align}
                         
By exploiting the fact that $q(\state_T|\state_{1:T-1})=\mathcal N\left(\state_T;\mu_T,\Sigma_T\right)$, we can derive closed-form solutions for 
\begin{align}
\state_T^* \;=\; \argmax_{\state_T \in \mathbb G}\; \mathcal N\left(\state_T;\mu_T,\Sigma_T\right) \label{eq:sTstar}
\end{align}
when $\mathbb G$ has special structure, which \emph{enables us to apply gradient descent to solve this constrained-optimization problem} (examples below). With a closed form solution to \eqref{eq:sTstar}, we can easily compute \eqref{eq:sAllstar} using \textit{unconstrained}-optimization as follows:
\begin{align}
 \state_{1:T}^* 
 &= \argmax_{\state_{1:T-1}\in \mathbb R^{2(T-1)}} \argmax_{\state_T \in \mathbb G_{\text{line-segment}}}\; q(\state_T|\state_{1:T-1})\prod_{t=1}^{T-1} q(\state_t|\state_{1:t-1}) \\
  \state_{1:T-1}^*   &= \underbrace{\argmax_{\state_{1:T-1}\in \mathbb R^{2(T-1)}}}_{\text{unconstrained optimization}} \underbrace{q(\state_T^*|\state_{1:t-1}) \prod_{t=1}^{T-1} q(\state_t|\state_{1:t-1}).}_{\text{objective function of }\state_{1:T-1}} \label{eq:s1Tm1star}
\end{align}

Note that \eqref{eq:s1Tm1star} only helps solve \eqref{eq:sAllstar} if \eqref{eq:sTstar} has a closed-form solution. We detail example of goal-sets with such closed-form solutions in the following subsections.
                         


\subsubsection{Point goal-set}

The solution to \eqref{eq:sTstar} in the case of a single desired goal $g\in\mathbb R^D$ is simply:
\begin{align}
\mathbb{G}_{\text{point}} &\;\doteq\; \{\goal_T\}, \\
\state_{T,\text{point}}^*
&\;\doteq\; \argmax_{\state_T \in \mathbb{G}_{\text{point}}} \mathcal N\left(\state_T;\mu_T,\Sigma_T\right) \nonumber \\
&\;=\; \goal_T.
\end{align}
More generally, multiple point goals help define \textit{optional} end points for planning: where the agent only need reach one valid end point (see Fig.~\ref{fig:finalstatevis} for examples), formulated as:
\begin{align}
\mathbb{G}_{\text{points}} &\;\doteq\; \{\goal_T^k\}_{k=1}^K, \\
\state_{T,\text{points}}^*
&\;\doteq\; \argmax_{\goal_T^k \in \mathbb{G}_{\text{points}}} \mathcal N\left(\goal_T^k;\mu_T,\Sigma_T\right). \label{eq:multipointgoalsetsoln}
\end{align}

\subsubsection{Line-segment goal-set}

We can form a goal-set as a finite-length line segment, connecting point $\mathbf a \in \mathbb{R}^D$ to point $\mathbf b \in \mathbb{R}^D$: 
\begin{align}
g_{\text{line}}(u) &\;\doteq\; \mathbf{a} + u\cdot(\mathbf{b}-\mathbf{a}),\; u \in \mathbb R, \\
\mathbb G_{\text{line-segment}}^{\mathbf{a}\rightarrow\mathbf{b}} &\;\doteq\; \{g_{\text{line}}(u): u \in [0,1]\}.
\end{align}
The solution to \eqref{eq:sTstar} in the case of line-segment goals is:
\begin{align}
\state_{T,\text{line-segment}}^*
&\;\doteq\; \argmax_{\state_T \in \mathbb G_{\text{line-segment}}^{\mathbf{a}\rightarrow\mathbf{b}}}\; \mathcal N\left(\state_T;\mu_T,\Sigma_T\right) \label{eq:sLineStarDef} \\
&\;=\; \mathbf a + \min\left(1,\; \max\left(0,\; \frac{(\mathbf{b}-\mathbf{a})^\top \Sigma^{-1}_T (\mu_T - \mathbf{a})}{(\mathbf{b}-\mathbf{a})^\top \Sigma^{-1}_T (\mathbf{b}-\mathbf{a})}\right)\right) \cdot (\mathbf b - \mathbf a).
\end{align}

\textbf{Proof:}

To solve \eqref{eq:sLineStarDef} is to find which point along the line $g_{\text{line}}(u)$ maximizes $\mathcal N\left(\cdot;\mu_T,\Sigma_T\right)$ subject to the constraint $0 \leq u \leq 1$:
\begin{align}
u^* 
&\;\doteq\; \argmax_{u \in [0,1]}\; \mathcal N\left(g_{\text{line}}(u);\mu_T,\Sigma_T\right)) \nonumber \\
&\;=\; \argmin_{u \in [0,1]}\; \underbrace{(g_{\text{line}}(u) - \mu_T)^{\top} \Sigma_T^{-1} (g_{\text{line}}(u) - \mu_T)}_{\mathcal{L}_u(u)}.
\end{align}
Since $\mathcal{L}_u$ is convex, the optimal value $u^*$ is value closest to the unconstrained $\argmax$ of $\mathcal{L}_u(u)$, subject to $0 \leq u \leq 1$:
\begin{align}
u_{\mathbb R}^* &\;\doteq\; \argmax_{u\in\mathbb R}\; \mathcal{L}_u(u), \\
u^* &\;=\; \argmin_{u \in [0,1]}\; \mathcal{L}_u(u) \nonumber \\
&\;=\; \min\left(1,\; \max\left(0,\; u_{\mathbb R}^*\right)\right).
\end{align}
We now solve for $u_{\mathbb R}^*$:
\begin{align}
u_{\mathbb R}^* = u : 0 = \deriv{\mathcal{L}(u)}{u} &= \deriv{\left((g_{\text{line}}(u) - \mu_T)^{\top} \Sigma_T^{-1} (g_{\text{line}}(u) - \mu_T)\right)}{u} \nonumber \\
  &= 2 \cdot \deriv{(g_{\text{line}}(u) - \mu_T)^{\top}}{u} \Sigma_T^{-1} (g_{\text{line}}(u) - \mu_T) \nonumber \\
  &= 2 \cdot \deriv{(\mathbf{a} + u\cdot(\mathbf{b}-\mathbf{a}) - \mu_T)^{\top}}{u} \Sigma_T^{-1} (\mathbf{a} + u\cdot(\mathbf{b}-\mathbf{a}) - \mu_T) \nonumber \\
  &= 2 \cdot (\mathbf{b}-\mathbf{a})^\top \Sigma_T^{-1} (\mathbf{a} + u\cdot(\mathbf{b}-\mathbf{a}) - \mu_T), \nonumber \\
u_{\mathbb R}^* &= \frac{(\mathbf{b}-\mathbf{a})^\top \Sigma_T^{-1} (\mu_T - \mathbf{a})}{(\mathbf{b}-\mathbf{a})^\top \Sigma_T^{-1} (\mathbf{b}-\mathbf{a})},
\end{align}
which gives us:
\begin{align}
\state_{T,\text{line-segment}}^*
&\;=\; g_{\text{line}}(u^*) \nonumber \\
&\;=\; \mathbf{a} + u^*\cdot(\mathbf{b}-\mathbf{a}) \nonumber \\
&\;=\; \mathbf{a} + \min\left(1,\; \max\left(0,\; u_{\mathbb R}^*\right)\right)\cdot(\mathbf{b}-\mathbf{a}) \nonumber \\
&\;=\; \mathbf a + \min\left(1,\; \max\left(0,\; \frac{(\mathbf{b}-\mathbf{a})^\top \Sigma^{-1}_T (\mu_T - \mathbf{a})}{(\mathbf{b}-\mathbf{a})^\top \Sigma^{-1}_T (\mathbf{b}-\mathbf{a})}\right)\right) \cdot (\mathbf b - \mathbf a).
\end{align}

\subsubsection{Multiple-line-segment goal-set:}\label{sec:multiLineSegmentGoalSet}

More generally, we can combine multiple line-segments to form piecewise linear ``paths'' we wish to follow.
By defining a path that connects points $(\mathbf{p}_0, \mathbf{p}_1,...,\mathbf{p}_N)$, we can evaluate $\mathcal{L}_u^i$ for each $\mathbb G_{\text{line-segment}}^{\mathbf{p}_i\rightarrow\mathbf{p}_{i+1}}$, select the optimal segment $i^*=\argmax_i\mathcal{L}_u^i$, and use the segment $i^*$'s solution to $u^*$ to compute $s^*_T$. Examples shown in Fig.~\ref{fig:linesegvis}.

\subsubsection{Polygon goal-set}

Instead of a route or path, a user (or program) may wish to provide a general \textit{region} the agent should go to, and state within that region being equally valid. Polygon regions (including both boundary and interior) offer closed form solution to \eqref{eq:sTstar} and are simple to specify. A polygon can be specified by an ordered sequence of vertices $(\mathbf{p}_0, \mathbf{p}_1,...,\mathbf{p}_N) \in\mathbb R^{N\times 2}$. Edges are then defined as the sequence of line-segments between successive vertices (and a final edge between first and last vertex): $\left((\mathbf{p}_0, \mathbf{p}_1),...,(\mathbf{p}_{N-1}, \mathbf{p}_N), (\mathbf{p}_N, \mathbf{p}_0)\right)$. Examples shown in Fig.~\ref{fig:regionvis} and \ref{fig:bigregionvis}. 

Solving \eqref{eq:sTstar} with a polygon has two cases: depending whether $\mu_T$ is \textit{inside} the polygon, or \textit{outside}. If $\mu_T$ lies inside the polygon, then the optimal value for $\state_T^*$ that maximizes $\mathcal N(\state_T^*; \mu_T, \Sigma_T)$ is simply $\mu_T$: the mode of the Gaussian distribution. Otherwise, if $\mu_T$ lies outside the polygon, then the optimal value $\state_T^*$ will lie on one of the polygon's edges, solved using \ref{sec:multiLineSegmentGoalSet}.
 
 \subsection{Waypointer Details} \label{app:smartwaypointer}

The waypointer uses the CARLA planner's provided route to generate waypoints. In the constrained-based planning goal likelihoods, we use this route to generate waypoints without interpolating between them. In the relaxed goal likelihoods, we interpolate this route to every $2$ meters, and use the first $20$ waypoints. As mentioned in the main text, one variant of our approach uses a ``smart'' waypointer. This waypointer simply removes nearby waypoints closer than $5$ meters from the vehicle when a green light is observed in the measurements provided by CARLA, to encourage the agent to move forward, and removes far waypoints beyond $5$ meters from the vehicle when a red light is observed in the measurements provided by CARLA. Note that the performance differences between our method without the smart waypointer and our method with the smart waypointer are small: the only signal in the metrics is that the smart waypointer improves the vehicle's ability to stop for red lights, however, it is quite adept at doing so without the smart waypointer.
 
\subsection{Constructing Goal Sets} \label{app:goalsetconstruction}
Given the in-lane waypoints generated by CARLA's route planner, we use these to create Point goal sets, Line-Segment goal sets, and Polygon Goal-Sets, which respectively correspond to the (A) Final-State Indicator, (B) Line-Segment Final-State Indicator, and (C) Final-State Region Indicator described in Section~\ref{sec:goallikelihoods}. For (A), we simply feed the waypoints directly into the Final-State Indicator, which results in a constrained optimization to ensure that $S_T \in \mathbb G = \{g_T^k\}_{k=1}^K$. We also included the vehicle's current position in the goal set, in order to allow it to stop. The gradient-descent based optimization is then formed from combining Eq.~\ref{eq:s1Tm1star} with Eq.~\ref{eq:multipointgoalsetsoln}. The gradient to the nearest goal of the final state of the partially-optimized plan encourage the optimization to move the plan closer to that goal. We used $K=10$. We applied the same procedure to generate the goal set for the (B) Line Segment indicator, as the waypoints returned by the planner are ordered. Finally, for the (C) Final-State Region Indicator (polygon), we used the ordered waypoints as the ``skeleton'' of a polygon that surrounds. It was created by adding a two vertices for each point $\mathbf{v}_t$ in the skeleton at a distance $1$ meter from $\mathbf{v}_t$ perpendicular to the segment connecting the surrounding points $(\mathbf{v}_{t-1},\mathbf{v}_{t+1})$. This resulted in a goal set $\mathbb G_{\text{polygon}} \supset \mathbb G_{\text{line-segment}}$, as it surrounds the line segments. The (F) Gaussian Final-State Mixture goal set was constructed in the same way as (A), and also used when the pothole costs were added.

 For the methods we implemented, the task is to drive the \emph{furthest} road location from the vehicle's initial position. Note that this protocol more difficult than the one used in prior work~\cite{codevilla2018end,liang2018cirl,sauer2018conditional,li2018rethinking,codevilla2019exploring}, which has no distance guarantees between start positions and goals, and often results in shorter paths. 

 \newlength{\goalvislength}
\setlength{\goalvislength}{.24\textwidth}

 \begin{figure}
 \begin{subfigure}[t]{\goalvislength}
    \includegraphics[width=\textwidth]{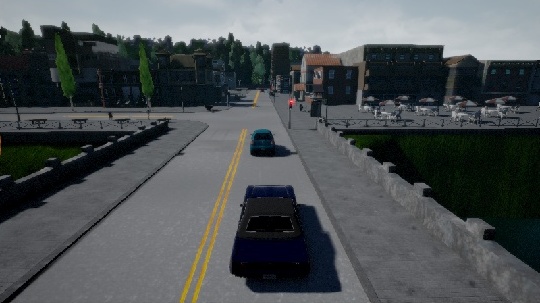}
\end{subfigure} 
 \begin{subfigure}[t]{\goalvislength}
    \includegraphics[width=\textwidth]{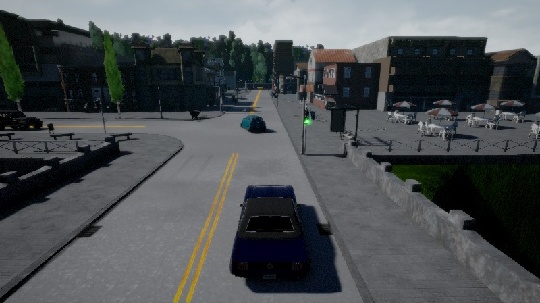}
\end{subfigure}
 \begin{subfigure}[t]{\goalvislength}
    \includegraphics[width=\textwidth]{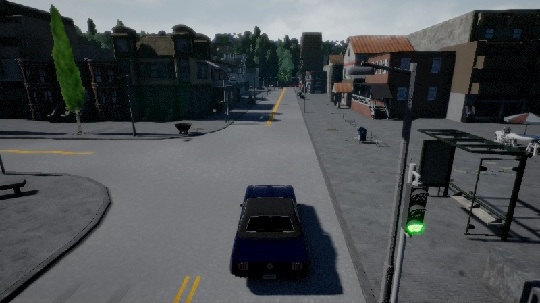}
\end{subfigure}
 \begin{subfigure}[t]{\goalvislength}
    \includegraphics[width=\textwidth]{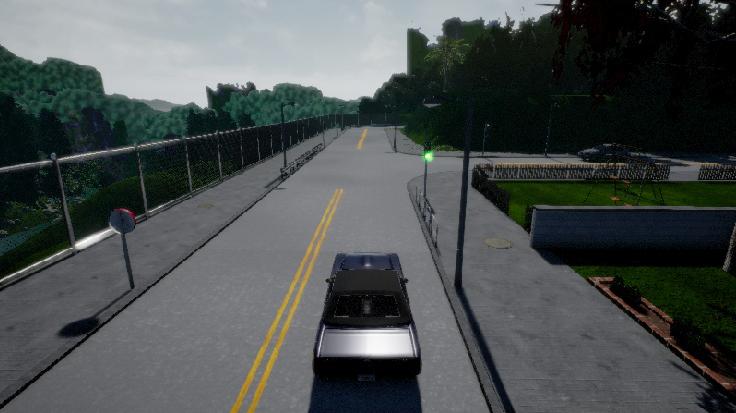}
\end{subfigure}
\\
 \begin{subfigure}[t]{\goalvislength}
    \includegraphics[angle=90,width=\textwidth]{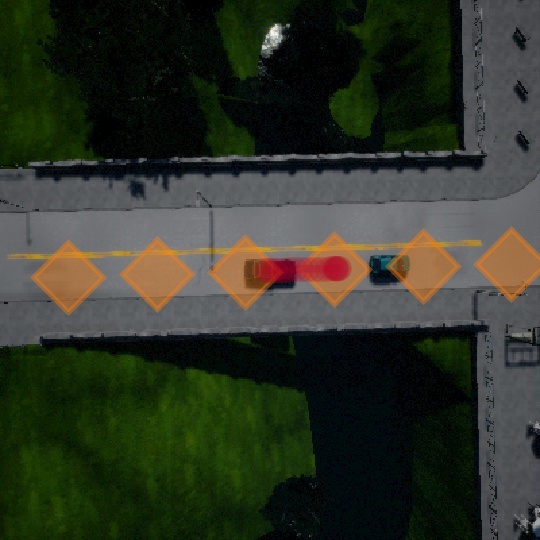}
\end{subfigure}
 \begin{subfigure}[t]{\goalvislength}
    \includegraphics[angle=90,width=\textwidth]{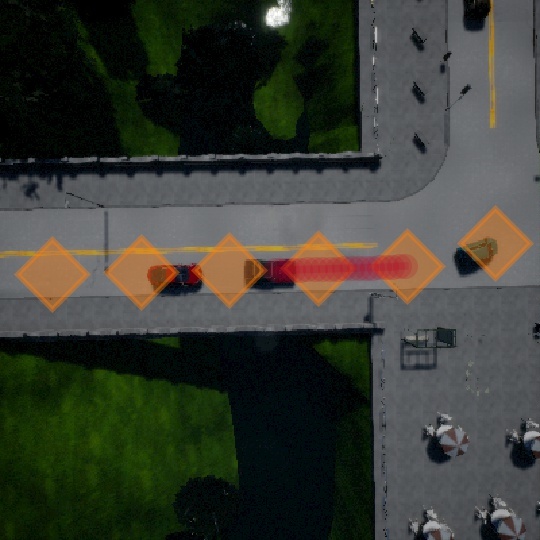}
\end{subfigure}
 \begin{subfigure}[t]{\goalvislength}
    \includegraphics[angle=90,width=\textwidth]{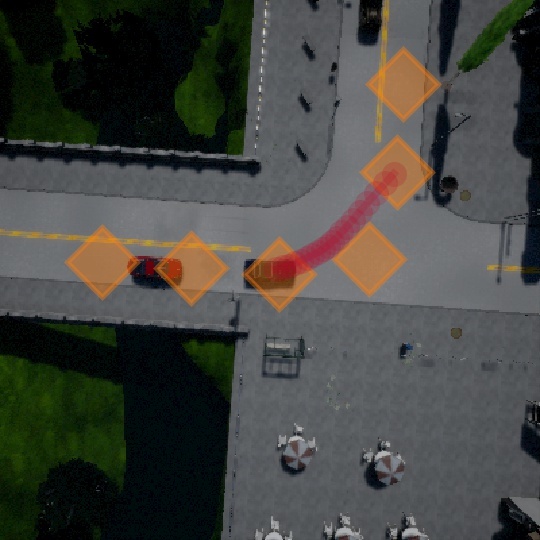}
\end{subfigure}
 \begin{subfigure}[t]{\goalvislength}
    \includegraphics[angle=90,width=\textwidth]{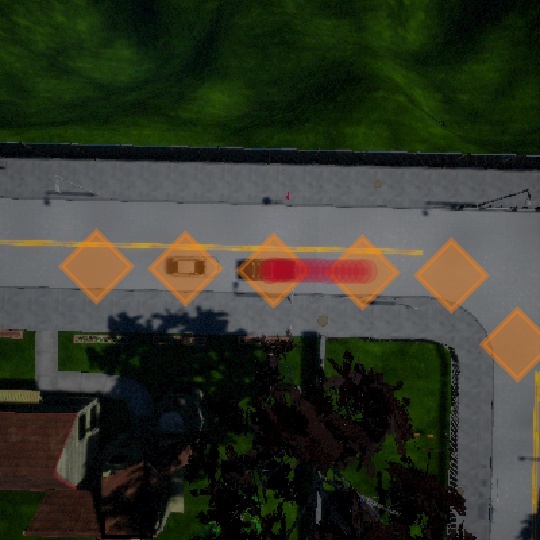}
\end{subfigure}
    \caption{Planning with the Final State Indicator yields plans that end at one of the provided locations. The orange diamonds indicate the locations in the goal set. The red circles indicate the chosen plan.} \label{fig:finalstatevis}
 \end{figure}

 \begin{figure}
 \begin{subfigure}[t]{\goalvislength}
    \includegraphics[width=\textwidth]{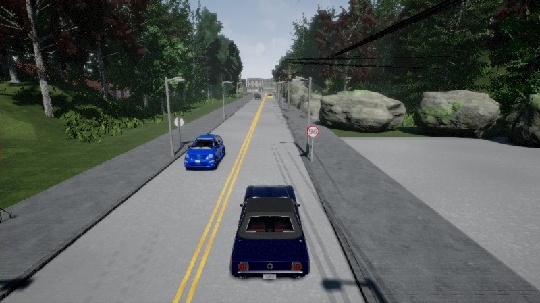}
\end{subfigure} 
 \begin{subfigure}[t]{\goalvislength}
    \includegraphics[width=\textwidth]{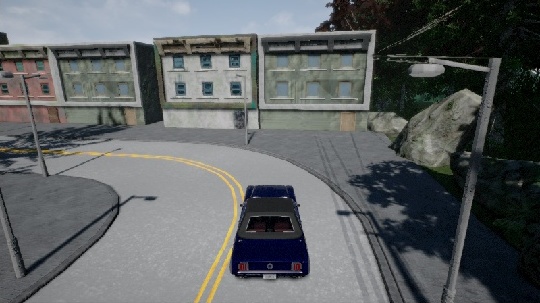}
\end{subfigure}
 \begin{subfigure}[t]{\goalvislength}
    \includegraphics[width=\textwidth]{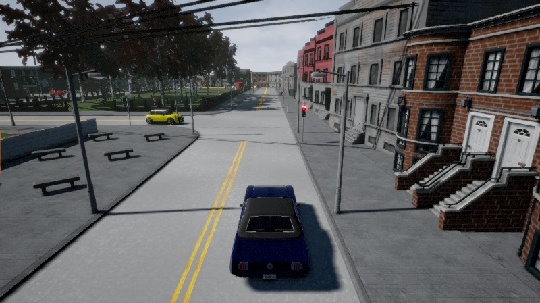}
\end{subfigure}
 \begin{subfigure}[t]{\goalvislength}
    \includegraphics[width=\textwidth]{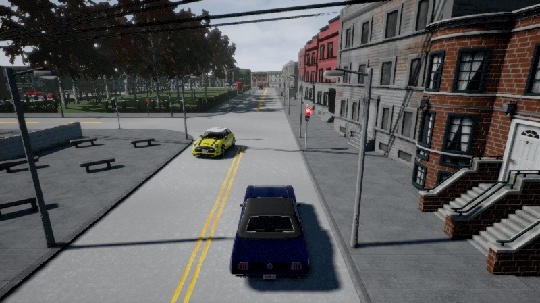}
\end{subfigure}
\\
 \begin{subfigure}[t]{\goalvislength}
    \includegraphics[angle=90,width=\textwidth]{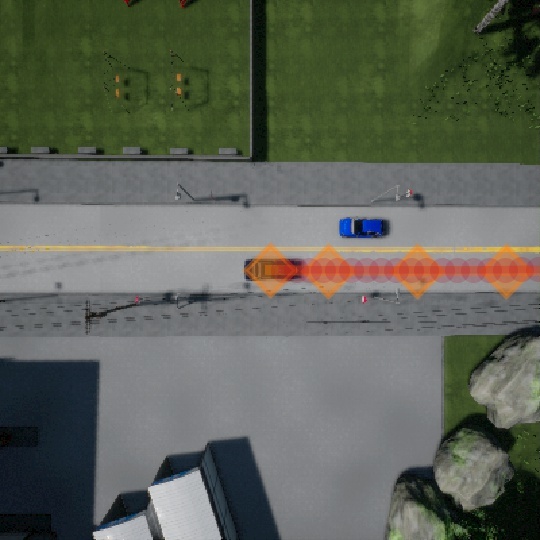}
\end{subfigure}
 \begin{subfigure}[t]{\goalvislength}
    \includegraphics[angle=90,width=\textwidth]{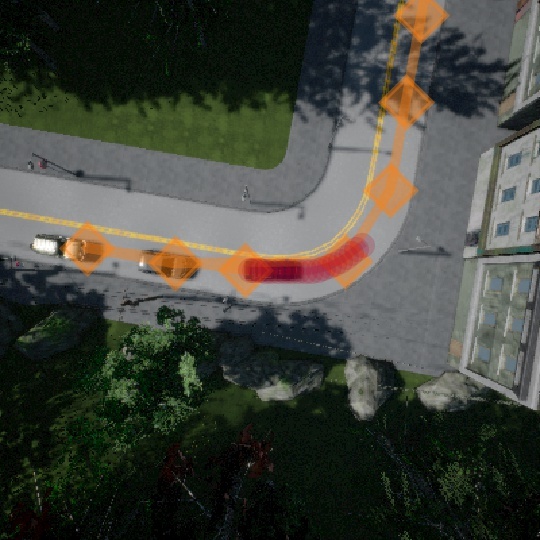}
\end{subfigure}
 \begin{subfigure}[t]{\goalvislength}
    \includegraphics[angle=90,width=\textwidth]{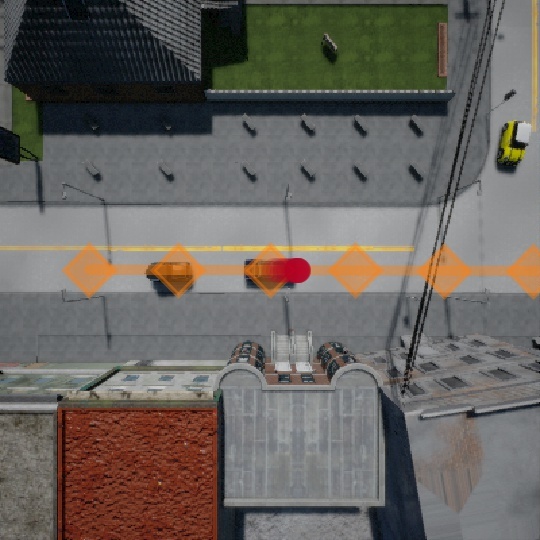}
\end{subfigure}
 \begin{subfigure}[t]{\goalvislength}
    \includegraphics[angle=90,width=\textwidth]{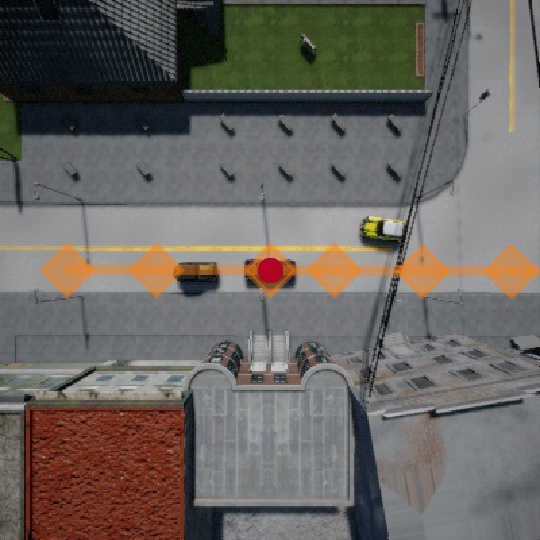}
\end{subfigure}
    \caption{Planning with the Line Segment Final State Indicator yields plans that end along one of the segments. The orange diamonds indicate the endpoints of each line segment. The red circles indicate the chosen plan.} \label{fig:linesegvis}
 \end{figure}
 

 \begin{figure}
 \begin{subfigure}[t]{\goalvislength}
    \includegraphics[width=\textwidth]{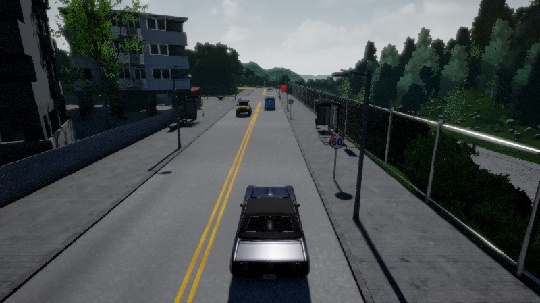}
\end{subfigure} 
 \begin{subfigure}[t]{\goalvislength}
    \includegraphics[width=\textwidth]{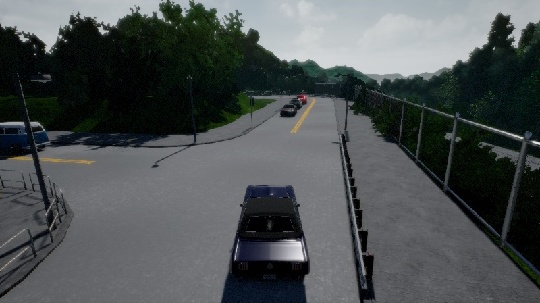}
\end{subfigure}
 \begin{subfigure}[t]{\goalvislength}
    \includegraphics[width=\textwidth]{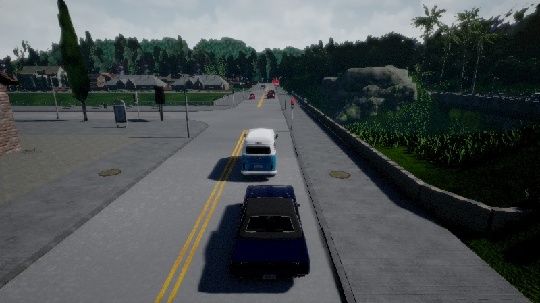}
\end{subfigure}
 \begin{subfigure}[t]{\goalvislength}
    \includegraphics[width=\textwidth]{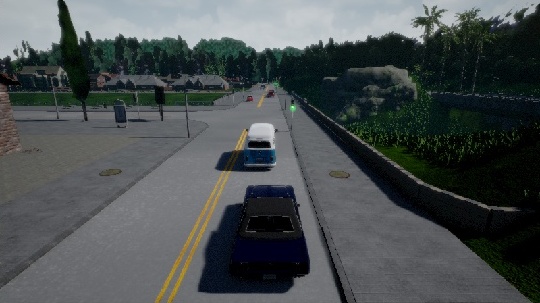}
\end{subfigure}
\\
 \begin{subfigure}[t]{\goalvislength}
    \includegraphics[angle=90,width=\textwidth]{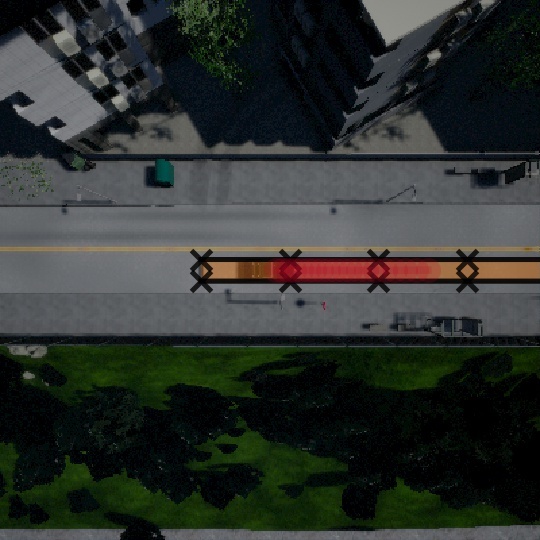}
\end{subfigure}
 \begin{subfigure}[t]{\goalvislength}
    \includegraphics[angle=90,width=\textwidth]{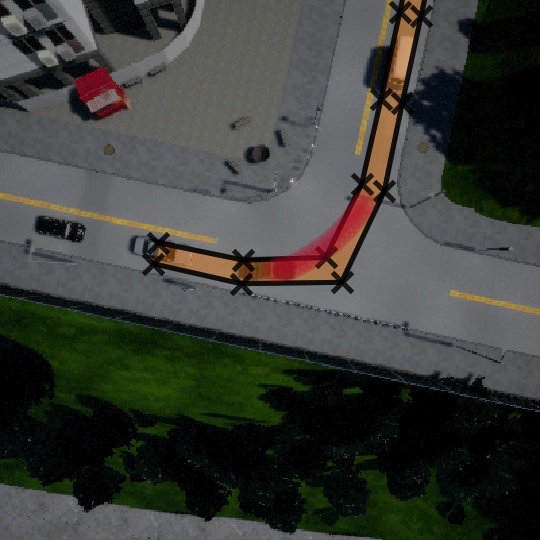}
\end{subfigure}
 \begin{subfigure}[t]{\goalvislength}
    \includegraphics[angle=90,width=\textwidth]{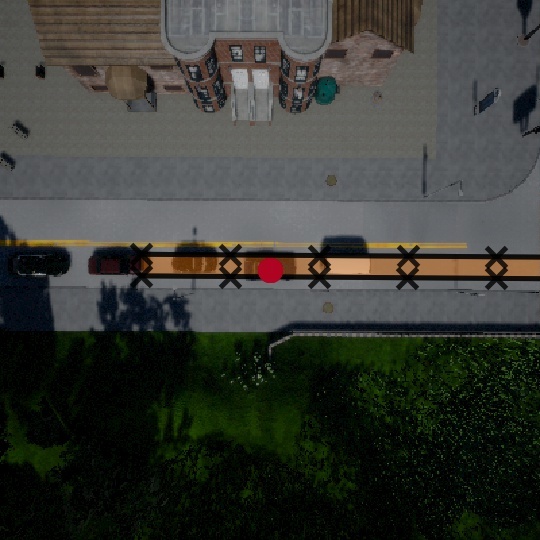}
\end{subfigure}
 \begin{subfigure}[t]{\goalvislength}
    \includegraphics[angle=90,width=\textwidth]{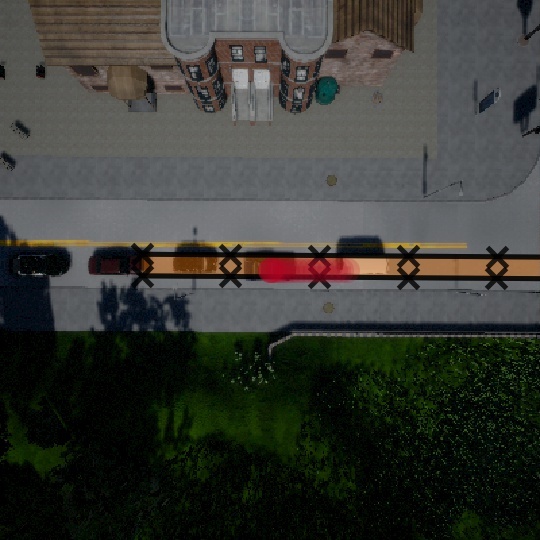}
\end{subfigure}
    \caption{Planning with the Region Final State Indicator yields plans that end inside the region. The orange polygon indicates the region. The red circles indicate the chosen plan.} \label{fig:regionvis}
 \end{figure}
 
  \begin{figure}
 \begin{subfigure}[t]{\goalvislength}
    \includegraphics[width=\textwidth]{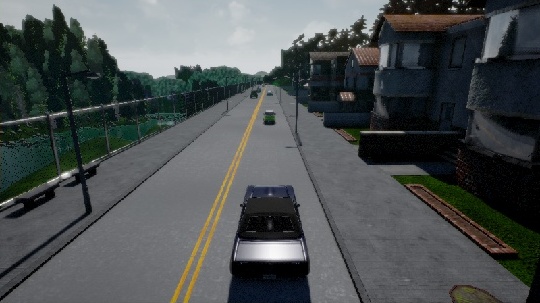}
\end{subfigure} 
 \begin{subfigure}[t]{\goalvislength}
    \includegraphics[width=\textwidth]{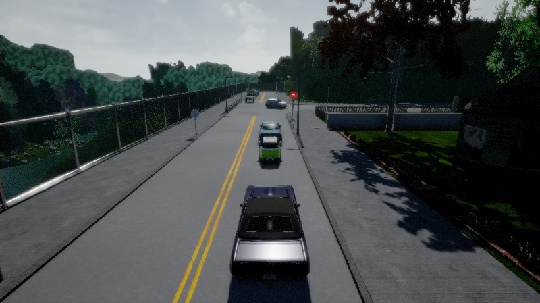}
\end{subfigure}
 \begin{subfigure}[t]{\goalvislength}
    \includegraphics[width=\textwidth]{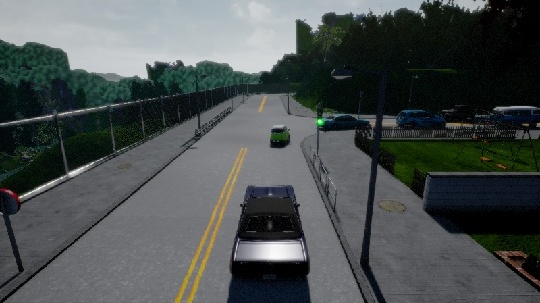}
\end{subfigure}
 \begin{subfigure}[t]{\goalvislength}
    \includegraphics[width=\textwidth]{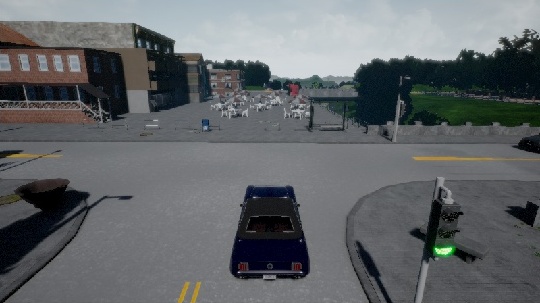}
\end{subfigure}
\\
 \begin{subfigure}[t]{\goalvislength}
    \includegraphics[angle=90,width=\textwidth]{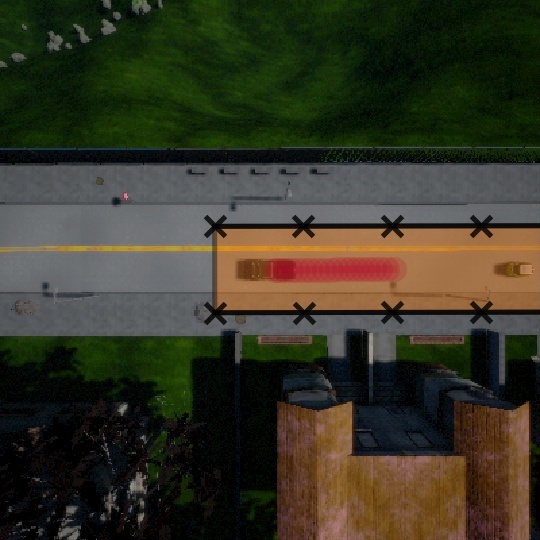}
\end{subfigure}
 \begin{subfigure}[t]{\goalvislength}
    \includegraphics[angle=90,width=\textwidth]{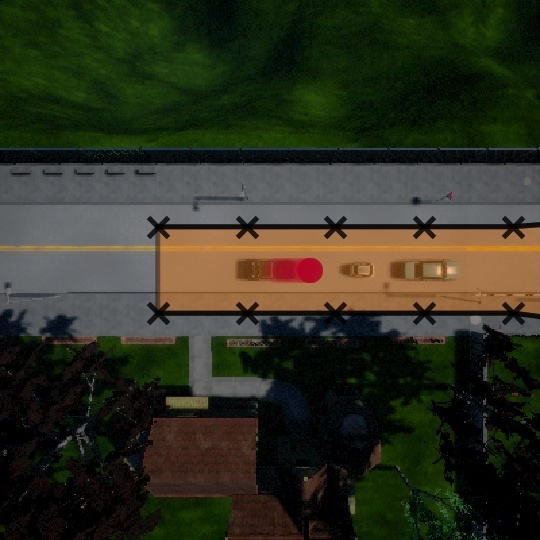}
\end{subfigure}
 \begin{subfigure}[t]{\goalvislength}
    \includegraphics[angle=90,width=\textwidth]{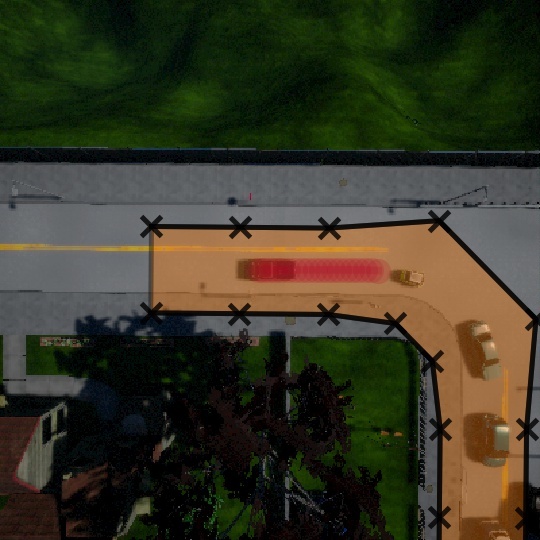}
\end{subfigure}
 \begin{subfigure}[t]{\goalvislength}
    \includegraphics[angle=90,width=\textwidth]{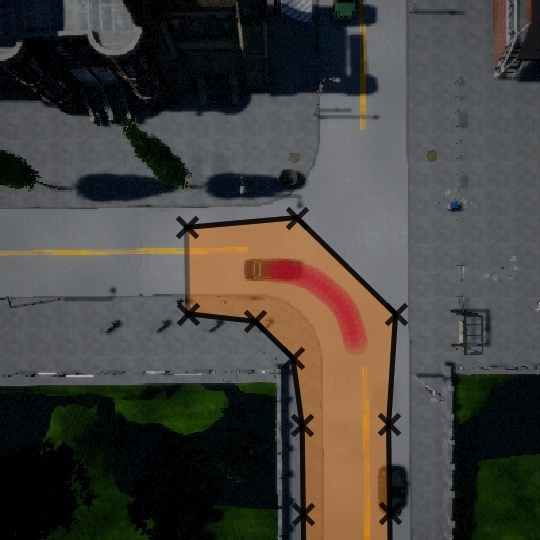}
\end{subfigure}
    \caption{Planning with the Region Final State Indicator yields plans that end inside the region. The orange polygon indicates the region. The red circles indicate the chosen plan. Note even with a wider goal region than Fig.~\ref{fig:regionvis}, the vehicle remains in its lane, due to the imitation prior. Despite their coarseness, these wide goal regions still provide useful guidance to the vehicle.} \label{fig:bigregionvis}
 \end{figure}

 \subsection{Planning visualizations} \label{app:goallikelihoodvis}
 Visualizations of examples of our method deployed with different goal likelihoods are shown in Fig.~\ref{fig:finalstatevis}, Fig.~\ref{fig:linesegvis}, Fig.~\ref{fig:regionvis}, and Fig.~\ref{fig:bigregionvis}.
 
\section{Architecture and Training Details} \label{app:architecture}
The architecture of $q(\State|\phi)$ is shown in Table~\ref{tab:archdetailed}. 
\begin{table*}[th]
\centering
\caption{Detailed Architecture that implements $\state_{1:T}=f(\latent_{1:T}, \phi)$. Typically, $T=40$, $D=2$,$H = W = 200$.}
\label{tab:archdetailed} 
\resizebox{.98\textwidth}{!}{
\begin{tabular}{lllllllll}
\toprule
Component & Input [dimensionality] & Layer or Operation  & Output [dimensionality] & Details \\
\midrule
\multicolumn{4}{l}{\emph{Static featurization of context:} $\phi=\{\chi,\state_{-\tau:0}^{1:A}\}$.} &\\
\midrule
MapFeat  & $\chi~[H,W,2]$ & 2D Convolution & ${}^{1}\chi~[H,W, 32]$ & $3\times3$ stride 1, ReLu \\
MapFeat  & ${}^{i-1}\chi~[H,W,32]$ & 2D Convolution & ${}^{i}\chi~[H, W, 32]$ & $3\times3$ stride 1, ReLu, $i \in [2, \dots, 8]$ \\
MapFeat  & ${}^{8}\chi~[H,W,32]$ & 2D Convolution & $\bGamma~[H, W, 8]$ & $3\times3$ stride 1, ReLu\\
PastRNN & $\state_{-\tau:0}~[\tau+1, D]$ & RNN & $\alpha~[32]$ & GRU across time dimension \\
\midrule
\multicolumn{4}{l}{\emph{Dynamic generation via loop:} $\mathrm{for}~t\in\{0,\dots,T-1\}$.}\\
\midrule
MapFeat & $\state_t~[D]$ & Interpolate & $\gamma_t=\bGamma(\state_t)~[8]$ & Differentiable interpolation\\
JointFeat & $\gamma_t,\state_0,{}^{2}\eta,\alpha,\blambda$ & $\gamma_t\oplus \state_0\oplus{}^{2}\eta\oplus\alpha\oplus\blambda$  & $\rho_t~[D+50+32+1]$ &  Concatenate ($\oplus$) \\
FutureRNN  & $\rho_t~[D+50+32+1]$ & RNN & ${}^{1}\rho_t~[50]$  & GRU\\
FutureMLP  & ${}^{1}\rho_t~[50]$ & Affine (FC) & ${}^{2}\rho_t~[200]$  & Tanh activation\\
FutureMLP  & ${}^{2}\rho_t[200]$ & Affine (FC) & $m_t~[D],\;\xi_t~[D,D]$  & Identity activation\\
MatrixExp & $\xi_t~[D,D]$& $\mathrm{expm}(\xi_t + \xi_t^{a,\mathrm{transpose}})$ & $\bsigma_t~[D,D]$ & Differentiable Matrix Exponential \cite{rhinehart2018r2p2} \\
VerletStep & $\state_{t},\state_{t-1},m_t,\bsigma_t,\latent_t$ & $2\state_{t}-\state_{t-1}+m_t+\bsigma_t\latent_t$ & $\state_{t+1}~[D]$ & \\
\bottomrule
\end{tabular}
}
\end{table*}

\subsection{Prior Visualization and Statistics} \label{app:prior_visualization}
\begin{figure}[h]
    \centering
        \begin{subfigure}[t]{.32\columnwidth}
        \includegraphics[width=\textwidth]{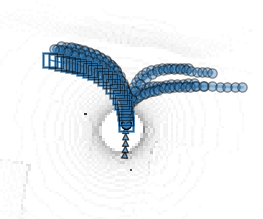}
    \end{subfigure}
    \begin{subfigure}[t]{.32\columnwidth}
     \includegraphics[width=\textwidth]{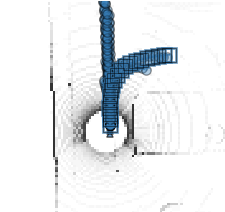}
    \end{subfigure}
        \caption{{\emph{Left}: Samples from the prior, $q(\State|\phi)$, go left or right. \emph{Right:} Samples go forward or right.}} \label{fig:samples}
\end{figure}

We show examples of the priors multimodality in Fig.~\ref{fig:samples}

\subsubsection{Statistics of Prior and Goal Likelihoods}
Following are the values of the planning criterion on $N\approx 8\cdot 10^3$ rounds from applying the ``Gaussian Final-State Mixture'' to Town01 Dynamic. Mean of $\log q(\state^*|\phi)\approx 104$. Mean of $\log p(\mathcal G|\state^*, \phi)=-4$. This illustrates that while the prior's value mostly dominates the values of the final plans, the Gaussian Final-State Goal Mixture likelihood has a moderate amount of influence on the value of the final plan.
 
\subsection{Dataset} \label{app:dataset}
Before training $q(\State|\phi)$, we ran CARLA's expert in the dynamic world setting of Town01 to collect a dataset of examples. We have prepared the dataset of collected data for public release upon publication.  We ran the autopilot in Town01 for over
900 episodes of 100 seconds each in the presence of 100
other vehicles, and recorded the trajectory of every vehicle and the autopilot’s LIDAR observation. We randomized
episodes to either train, validation, or test sets. We created
sets of 60,701 train, 7586 validation, and 7567 test scenes,
each with 2 seconds of past and 4 seconds of future position
information at 10Hz. The dataset also includes 100 episodes
obtained by following the same procedure in Town02. 

\section{Baseline Details} \label{app:baselines}

\subsection{Conditional Imitation Learning of States (CILS):} We designed a conditional imitation learning baselines that predicts the setpoint for the PID-controller. Each receives the same scene observations (LIDAR) and is trained with the same set of trajectories as our main method. It uses nearly the same architecture as that of the original CIL, except it outputs setpoints instead of controls, and also observes the traffic light information. We found it very effective for stable control on straightaways. When the model encounters corners, however, prediction is more difficult, as in order to successfully avoid the curbs, the model must implicitly plan a safe path. We found that using the traffic light information allowed it to stop more frequently.
\subsection{Model-Based Reinforcement Learning:} 

\noindent {\bf Static-world}
To compare against a purely model-based reinforcement learning algorithm, we propose a model-based reinforcement learning baseline. This baseline first learns a forwards dynamics model $\state_{t+1}=f(\state_{t-3:t}, \action_{t})$ given observed expert data ($a_t$ are recorded vehicle actions). We use an MLP with two hidden layers, each 100 units. Note that our forwards dynamics model does not imitate the expert preferred actions, but only models what is physically possible. Together with the same LIDAR map $\bchi$ our method uses to locate obstacles, this baseline uses its dynamics model to plan a reachability tree \cite{lavalle2006planning} through the free-space to the waypoint while avoiding obstacles. The planner opts for the lowest-cost path that ends near the goal $C(\state_{1:T}; \goal_T)=||\state_T-\goal_T||_2 + \sum_{t=1}^T c(\state_t)$, where cost of a position is determined by $c(\state_t)=1.5\mathbbm{1}(\state_t < 1~\text{meters from any  obstacle})+0.75\mathbbm{1}(1 <= \state_t < 2~\text{meters from any obstacle}) + \dddot{\state_t\hspace{0pt}}$.

We plan forwards over 20 time steps using a breadth-first search search over CARLA steering angle $\{-0.3, -0.1, 0., 0.1, 0.3\}$, noting valid steering angles are normalized to $[-1, 1]$, with constant throttle at 0.5, noting the valid throttle range is $[0, 1]$. Our search expands each state node by the available actions and retains the 50 closest nodes to the waypoint. The planned trajectory efficiently reaches the waypoint, and can successfully plan around perceived obstacles to avoid getting stuck. To convert the LIDAR images into obstacle maps, we expanded all obstacles by the approximate radius of the car, 1.5 meters.

\noindent{\bf Dynamic-world}
We use the same setup as the Static-MBRL method, except we add a discrete temporal dimension to the search space (one $\mathbb R^2$ spatial dimension per T time steps into the future). All static obstacles remain static, however all LIDAR points that were known to collide with a vehicle are now removed: and replaced at every time step using a constant velocity model of that vehicle. We found that the main failure mode was due to both to inaccuracy in constant velocity prediction as well as the model's inability to perceive lanes in the LIDAR. The vehicle would sometimes wander into the opposing traffic's lane, having failed to anticipate an oncoming vehicle blocking its path.

\section{Robustness Experiments details} \label{app:robustness}
\subsection{Decoy Waypoints} 
\begin{figure}[tbh]
    \centering
    \begin{subfigure}[t]{\roblength}
       \begin{overpic}[width=\textwidth]{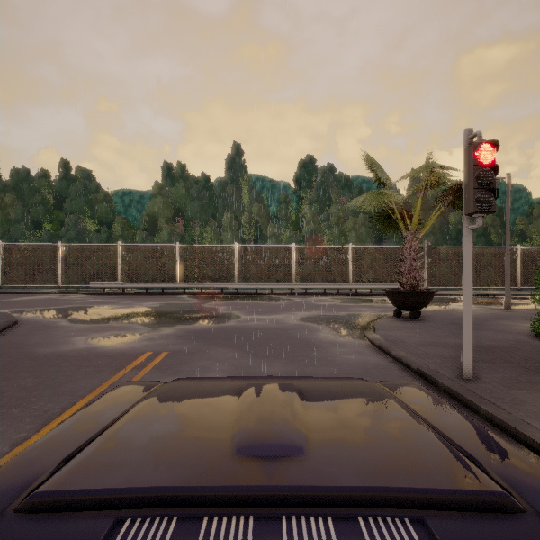} 
       \end{overpic}
    \end{subfigure}
    \begin{subfigure}[t]{\roblength}
       \begin{overpic}[width=\textwidth]{img/decoy_waypoints/plot_00000586_q1.png} 
       \end{overpic}
    \end{subfigure}
          \begin{subfigure}[t]{\roblength}
       \begin{overpic}[width=\textwidth]{img/wrongside/plot_00000058_q1.png}
       \end{overpic}
    \end{subfigure}
              \begin{subfigure}[t]{\roblength}
       \begin{overpic}[width=\textwidth]{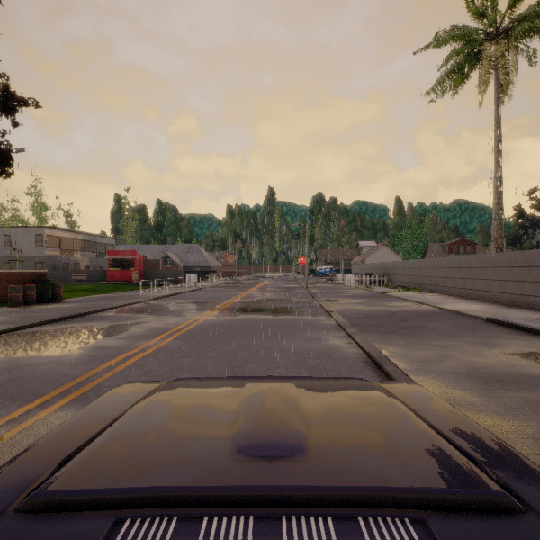}
       \end{overpic}
    \end{subfigure}
    
            \begin{subfigure}[t]{\roblength}
       \FBox{ 
       \begin{overpic}[width=\textwidth,angle=90]{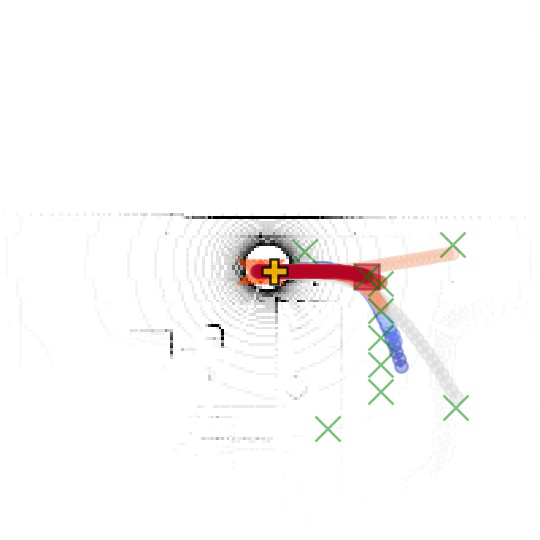} 
        \put(0,15){\includegraphics[width=0.7\textwidth]{img/legends/teaser_plotlegend.png}}
      \put(0,0){\includegraphics[width=0.7\textwidth]{img/legends/coolwarm_lth.png}}
       \end{overpic}
       }
          \end{subfigure}
    \begin{subfigure}[t]{\roblength}
       \FBox{ 
       \begin{overpic}[width=\textwidth,angle=90]{img/decoy_waypoints/plot_00000067_q0.png} 
       \end{overpic}
       }
    \end{subfigure}
    \FBox{ 
      \begin{subfigure}[t]{\roblength}
       \begin{overpic}[width=\textwidth,angle=90]{img/wrongside/plot_00000058_q0.png}
       \end{overpic}
    \end{subfigure} 
    }
    \FBox{ 
      \begin{subfigure}[t]{\roblength}
       \begin{overpic}[width=\textwidth,angle=90]{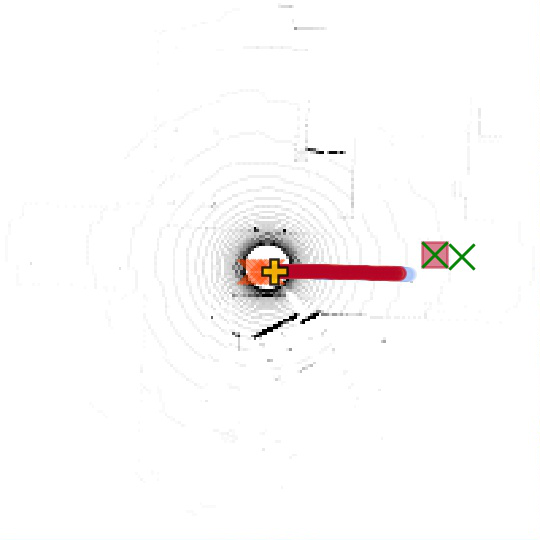}
       
       \end{overpic}
    \end{subfigure} 
    }
    \caption{Tolerating bad waypoints. The planner prefers waypoints in the distribution of expert behavior (on the road at a reasonable distance). \emph{Columns 1,2}: Planning with $\nicefrac{1}{2}$ decoy waypoints. \emph{Columns 3,4}: Planning with all waypoints on the wrong side of the road.} \label{fig:decoy-waypoints-appendix}
\end{figure}

In the decoy waypoints experiment, the perturbation distribution is $\mathcal N(0, \sigma=8m)$: each waypoint is perturbed with a standard deviation of $8$ meters.  One failure mode of this approach is when decoy waypoints lie on a valid off-route path at intersections, which temporarily confuses the planner about the best route. Additional visualizations are shown in Fig.~\ref{fig:decoy-waypoints-appendix}.

 \subsection{Plan Reliability Estimation} \label{app:planreliable}
Besides using our model to make a best-effort attempt to reach a user-specified goal, the fact that our model produces explicit likelihoods can also be leveraged to test the \emph{reliability} of a plan by evaluating whether reaching particular waypoints will result in human-like behavior or not. This capability can be quite important for real-world safety-critical applications, such as autonomous driving, and can be used to build a degree of fault tolerance into the system. We designed a classification experiment to evaluate how well our model can recognize safe and unsafe plans. We planned our model to  known good waypoints (where the expert actually went) and known bad waypoints (off-road) on 1650 held-out test scenes. We used the planning criterion to classify these as good and bad plans and found that we can detect these bad plans with $97.5\%$ recall and $90.2\%$ precision. This result indicates imitative models could be effective in estimating the reliability of plans. 
 
We determined a threshold on the planning criterion by single-goal planning to the expert’s final location on offline validation data and setting it to the criterion's mean minus one stddev. Although a more intelligent calibration could be performed by analyzing the information retrieval statistics on the offline validation, we found this simple calibration to yield reasonably good performance. We used 1650 test scenes to perform classification of plans to three different types of waypoints 1) where the expert actually arrived at time $T$ ($89.4\%$ reliable), 2) waypoints $20\mathrm{m}$ ahead along the waypointer-provided route, which are often near where the expert arrives ($73.8\%$ reliable) 3) the same waypoints from 2), shifted $2.5\mathrm{m}$ off of the road ($2.5\%$ reliable). This shows that our learned model exhibits a strong preference for valid waypoints. Therefore, a waypointer that provides expert waypoints via 1) half of the time, and slightly out-of-distribution waypoints via 3) in the other half, an ``unreliable'' plan classifier achieves $97.5\%$ recall and $90.2\%$ precision. 

\section{Pothole Experiment Details} \label{sec:potholedetails}
We simulated potholes in the environment by randomly inserting them in the cost map near each waypoint $i$ with offsets
distributed $\mathcal N_i(\mu\!\!=\!\![-15\mathrm{m}, 0\mathrm{m}], \Sigma=\mathrm{diag}([1,0.01]))$,  (i.e. mean-centered on the right side of the lane $15\mathrm{m}$ before each waypoint). We inserted pixels of root cost $-1e3$ in the cost map at a single sample of each $\mathcal N_i$, binary-dilated the cost map by $\nicefrac{1}{3}$ of the lane-width (spreading the cost to neighboring pixels), and then blurred the cost map by convolving with a normalized truncated Gaussian filter of $\sigma=1$ and truncation width $1$.

\section{Baseline Visualizations}
\begin{figure}[h]
    \centering
        \begin{subfigure}[t]{.32\columnwidth}
       \FBox{\includegraphics[width=\columnwidth,angle=90]{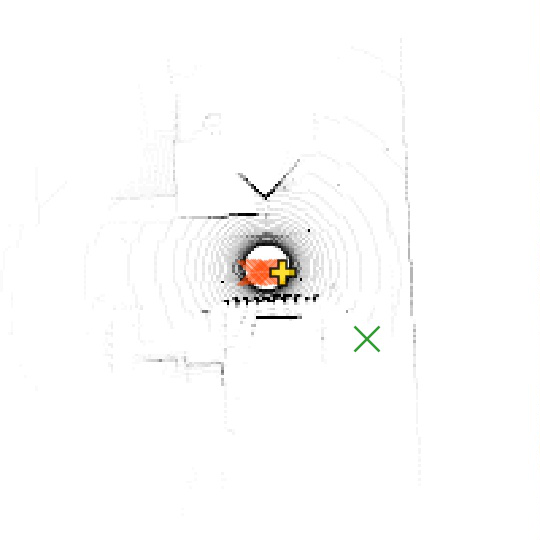}}
    \end{subfigure}
    \begin{subfigure}[t]{.32\columnwidth}
    \FBox{   \includegraphics[width=\columnwidth,angle=90]{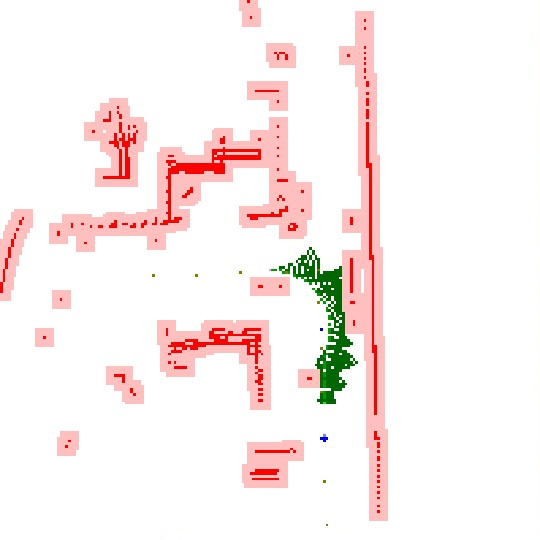} }
    \end{subfigure}
    \caption{Baseline methods we compare against.  The red crosses indicate the past 10 positions of the agent.     {\emph{Left: Imitation Learning baseline}: the green cross indicates the provided goal, and the yellow plus indicates the predicted setpoint for the controller.}         {\emph{Right: Model-based RL baseline:} the green regions indicate the model's predicted reachability, the red regions are post-processed LIDAR used to create its obstacle map.}}
    \label{fig:modelbased_comparison}
\end{figure}

See Fig.~\ref{fig:modelbased_comparison} for a visualization of our baseline methods. 

\section{Hyperparameters}
In order to tune the $\epsilon$ hyperparameter of the unconstrained likelihoods, we undertook the following binary-search procedure. When the prior frequently overwhelmed the posterior, we set $\epsilon\leftarrow 0.2\epsilon$, to yield tighter covariances, and thus \emph{more} penalty for failing to satisfy the goals. When the posterior frequently overwhelmed the prior, we set $\epsilon\leftarrow 5\epsilon$, to yield looser covariances, and thus \emph{less} penalty for failing to satisfy the goals. We executed this process three times: once for the ``Gaussian Final-State Mixture'' experiments (Section~\ref{sec:experiments}), once for the ``Noise Robustness'' Experiments (Section~\ref{sec:noise_robustness}), and once for the pothole-planning experiments (Section~\ref{sec:pothole_experiments}). Note that for the Constrained-Goal Likelihoods introduced no hyperparameters to tune.


\end{document}